\PassOptionsToPackage{table}{xcolor}
\documentclass{article}

\usepackage{iclr2027_preprint,times}

\usepackage[utf8]{inputenc}
\usepackage[T1]{fontenc}
\usepackage{booktabs}
\usepackage{array}
\usepackage{amsfonts}
\usepackage{amsmath}
\usepackage{amssymb}
\usepackage{amsthm}
\usepackage{nicefrac}
\usepackage{microtype}
\usepackage{graphicx}
\usepackage{multirow}
\usepackage{enumitem}
\usepackage{makecell}
\usepackage{xcolor}
\usepackage{etoolbox}
\usepackage{url}
\usepackage{hyperref}
\hypersetup{
  hidelinks,
  pdftitle={JUMP: Efficient Membership Inference on Fine-Tuned Diffusion Language Models},
  pdfauthor={Yeachan Jun and Albert No}
}
\usepackage{wrapfig}
\usepackage{subcaption}
\usepackage{placeins}
\usepackage{float}
\usepackage{needspace}
\definecolor{lightgrayrow}{RGB}{247,247,247}
\definecolor{oursrow}{RGB}{235,242,250}

\usepackage[capitalise,noabbrev]{cleveref}
\crefname{figure}{Figure}{Figures}
\Crefname{figure}{Figure}{Figures}

\crefname{equation}{Eq.}{Eqs.}
\Crefname{equation}{Eq.}{Eqs.}

\crefname{appendix}{Appendix}{Appendices}
\Crefname{appendix}{Appendix}{Appendices}
\graphicspath{{figures/}}

\newcommand{\dataD}{\mathcal{D}}
\newcommand{\Mtgt}{M_{\mathrm{tgt}}}
\newcommand{\Mref}{M_{\mathrm{ref}}}
\newcommand{\Mbase}{M_{\mathrm{base}}}
\newcommand{\mask}{\texttt{[MASK]}}
\newcommand{\method}{\textsc{JUMP}}
\newcommand{\methodmath}{\mathrm{JUMP}}
\newcommand{\methodfull}{\method{} (\emph{Joint Uncertainty-Guided Mask Probing})}

\title{JUMP: Efficient Membership Inference on Fine-Tuned Diffusion Language Models}

\author{%
  Yeachan Jun \\
  Department of Artificial Intelligence \\
  Yonsei University, Seoul, Korea \\
  \texttt{dpcks8942@yonsei.ac.kr}
  \And
  Albert No\thanks{Corresponding author.} \\
  Department of Artificial Intelligence \\
  Yonsei University, Seoul, Korea \\
  \texttt{albertno@yonsei.ac.kr}
}

\iclrfinalcopy

\begin{document}
\raggedbottom

\maketitle
\begin{abstract}
Membership inference attacks (MIAs) test whether a candidate example was used to train a language model.
Existing attacks on fine-tuned discrete diffusion language models (dLLMs) often aggregate reconstruction signals across many mask configurations, requiring repeated model evaluations.
We propose \methodfull{}, an efficient MIA that exploits the ability of dLLMs to predict masked tokens in parallel.
Using the pre-fine-tuning checkpoint as a reference, \method{} selects low-confidence positions, masks them jointly, and aggregates clipped target--reference reconstruction gaps.
This focuses the attack on positions that reveal stronger membership signals from fine-tuning.
After mask selection, all selected tokens are evaluated with one scoring query per model.
Across six MIMIR domains, \method{} improves mean ROC-AUC over a prior multi-mask attack from $0.819$ to $0.902$ on LLaDA and from $0.851$ to $0.942$ on Dream.
Including mask selection, it requires only three forward passes per example, compared with 32 for the baseline.
We further extend \method{} to the target-only setting by replacing target--reference scoring with \emph{relative token preference}, which compares the observed token with alternative predictions at the same masked position.
Target-Only \method{} achieves mean ROC-AUCs of $0.609$ and $0.638$ on LLaDA and Dream.
\end{abstract}
 
\section{Introduction} \label{sec:intro}

Membership inference attacks (MIAs) aim to determine whether a candidate example was included in a model's training data~\citep{shokri2017membership,yeom2018privacy}.  For large language models (LLMs), they provide a way to audit privacy risks associated with text used during pretraining or fine-tuning~\citep{carlini2021extracting,duan2024membership}.  Most existing LLM MIAs use autoregressive (AR) next-token probabilities as membership evidence, refining sequence likelihood through reference calibration, token selection, information-theoretic normalization, or comparisons with perturbed inputs~\citep{shi2024detecting,zhang2025minkpp,xie2024recall,mattern2023membership,chang2025context,tao2026informationtheoretic}.  Because each token is evaluated under a fixed left-to-right context, these methods mainly differ in how they select, calibrate, and aggregate a predetermined set of token scores.

Recent discrete diffusion language models (dLLMs), including LLaDA and Dream, offer a different interface for membership inference~\citep{nie2025large,ye2025dream}.  They generate text by reconstructing masked tokens from bidirectional context rather than predicting tokens strictly from left to right.  Two properties are particularly relevant.  First, an attacker can choose which tokens to mask and thereby change the reconstruction task; we call this \emph{any-order decodability}.  Second, one forward pass returns predictions for all masked positions in parallel; we call this \emph{parallel decodability}.  Together, these properties let the attacker control both the probed positions and their reconstruction context, making the mask set itself a central component of attack design.

Existing dLLM attacks collect membership evidence across multiple masked reconstructions.
SAMA~\citep{chen2026membership} targets fine-tuned dLLMs by comparing reconstruction losses between the target and a reference model, ideally the pre-fine-tuning checkpoint, across progressively masked inputs.
Other work has explored distributional statistics such as reconstruction-gain skewness~\citep{zhai2026membership}.
While these approaches demonstrate the value of masked reconstruction for membership inference, repeated probing increases cost, and random mask positions can include tokens with little membership evidence.
This motivates our central question: \emph{can informative positions be identified first and then scored together with a single masked query per model?}

In this work, we first consider membership in a dLLM's fine-tuning set when both the target model and its pre-fine-tuning checkpoint are available.
We propose \methodfull{}, which combines uncertainty-guided mask selection with single-pass scoring.
Our diagnostics show that low-reference-confidence positions exhibit stronger member--non-member separation in the reconstruction improvement induced by fine-tuning.
\method{} identifies these positions with a reference-based selector, masks them jointly, and aggregates clipped differences between the target and reference reconstruction log-probabilities.
The reference comparison controls for pre-existing predictability, while clipping limits the influence of extreme token scores.
Because all selected tokens are reconstructed in parallel, scoring requires only one masked query to each model.
Across six MIMIR domains~\citep{duan2024membership}, \method{} improves mean ROC-AUC over SAMA from $0.819$ to $0.902$ on LLaDA and from $0.851$ to $0.942$ on Dream.
Including the selector pass, it uses three forward passes per example rather than 32 in our SAMA configuration.

We further extend \method{} to the target-only setting, where the target's full predictive distribution serves as an internal comparison in place of a reference model.
Target-Only \method{} compares each observed token with the target model's own alternative predictions, filters near-deterministic positions, and estimates relative token preference through balanced grouped masking.
Across the same six MIMIR domains, it achieves mean ROC-AUCs of $0.609$ on LLaDA and $0.638$ on Dream, showing that target predictions alone retain a weaker but measurable membership signal.

\section{Related Work}
\label{sec:related}

\paragraph{Membership inference attacks.}
Given a target model $\Mtgt$ trained on $\dataD_{\mathrm{train}}$ and a candidate example $x$, an MIA asks whether $x \in \dataD_{\mathrm{train}}$~\citep{shokri2017membership,yeom2018privacy}. We write a score-based MIA as a scalar statistic
\begin{equation*}
  g(x;\,\Mtgt,\,\Mref) \in \mathbb{R},
\end{equation*}
where $\Mref$ may be absent for target-only attacks.
The attacker predicts membership when $g(x;\Mtgt,\Mref) > \eta$, and varying $\eta$ yields the ROC curve.
A classical example is the negative-loss statistic $g_{\mathrm{loss}}(x;\Mtgt)=-\ell(\Mtgt,x)$, which connects membership leakage to the generalization gap~\citep{yeom2018privacy}.
Reference-based attacks instead compare the target with a reference,
e.g., $g_{\mathrm{ref}}(x)= -\ell(\Mtgt,x)+\ell(\Mref,x)$, to reduce the confounding effect of intrinsic example difficulty~\citep{watson2022on,carlini2022membership,zarifzadeh2023low}.
Complementary work studies training-data extraction and privacy auditing~\citep{long2018understanding,carlini2019secret,carlini2021extracting,carlini2023quantifying,nasr2025scalable,steinke2023privacy,jagielski2023measuring,lukas2023analyzing,mireshghallah2022quantifying}.

\paragraph{MIA for autoregressive language models.}
For an autoregressive language model, next-token log-likelihood provides a natural membership score:
\begin{equation}
g_{\mathrm{AR}}(x)
=
\log p_{\mathrm{AR}}(x)
=
\sum_{i=1}^{L}\log p(x_i\mid x_{<i}).
\end{equation}
Methods such as Min-K\% and Min-K\%++ focus on low-probability tokens rather than the full sequence~\citep{shi2024detecting,zhang2025minkpp}, while ReCaLL, neighborhood-based attacks, and context-aware methods calibrate likelihood using alternative contexts or comparisons~\citep{xie2024recall,mattern2023membership,chang2025context}.
Recent work develops information-theoretic normalization of the membership signal~\citep{tao2026informationtheoretic}.
Although these attacks aggregate evidence differently, the conditional context for each token remains fixed by autoregressive decoding.

\paragraph{Discrete diffusion language models.}
Masked diffusion language models~\citep{austin2021structured,hoogeboom2021argmax,lou2024sedd,sahoo2024mdlm,shi2024simplified,ou2025radd}, including recent large-scale dLLMs such as LLaDA and Dream~\citep{nie2025large,ye2025dream}, define reconstruction distributions over arbitrary masked subsets.
Let $x=(x_1,\dots,x_L)$ and let $x_{\setminus S}$ denote the sequence where positions in $S\subseteq\{1,\dots,L\}$ are replaced by \mask.
A dLLM models $p_\theta(x_i \mid x_{\setminus S})$ for $i \in S$
and is trained with a masked reconstruction objective of the form
\begin{equation*}
  \mathcal{L}(\theta)
  = -\,\mathbb{E}_{x,\lambda,S}\!\left[
      \frac{1}{|S|}\sum_{i\in S}\log p_\theta(x_i \mid x_{\setminus S})
    \right],
  \qquad \lambda \sim U(0,1).
\end{equation*}
The attacker-facing consequence is that $S$ is not fixed.
Any-order decodability allows the attacker to choose which reconstruction task to query, and parallel decodability returns all $|S|$ masked-token distributions from one forward pass~\citep{uria2014deep,germain2015made,yang2019xlnet,hoogeboom2022autoregressive,ghazvininejad2019maskpredict,chang2022maskgit}.

\paragraph{MIA for diffusion language models.}
SAMA~\citep{chen2026membership}, the first dLLM-specific MIA, studies fine-tuning membership by comparing a fine-tuned target with its pre-fine-tuning checkpoint.
For each candidate sequence, it samples $T$ random mask sets $S_1,\dots,S_T$; for each $S_t$, the masked positions are reconstructed by both models, and their true-token log-probability gaps are averaged:
\begin{equation}
\label{eq:sama}
  g_{\mathrm{SAMA}}(x)
  = \frac{1}{T}\sum_{t=1}^{T}\frac{1}{|S_t|}\sum_{i\in S_t}
    \left[
      \log p_{\Mtgt}(x_i \mid x_{\setminus S_t})
      -
      \log p_{\Mref}(x_i \mid x_{\setminus S_t})
    \right].
\end{equation}
Q-Skew~\citep{zhai2026membership} takes a different approach, characterizing token-level memorization asymmetry across diffusion states with a quantile-weighted skewness statistic.
Both methods aggregate membership evidence over multiple masking conditions and therefore require repeated model evaluations.
In contrast, \method{} identifies informative positions first and scores them jointly, reducing the number of model evaluations.

\section{Threat Model}
\label{sec:threat}

\paragraph{Setting and access.}
We consider fine-tuning membership inference for a target dLLM $\Mtgt$ adapted on a private set $\dataD_{\mathrm{ft}}$.
Given a candidate sequence $x=(x_1,\ldots,x_L)$, the attacker determines whether $x\in\dataD_{\mathrm{ft}}$.
We assume grey-box score access: for any chosen mask set $S$, the attacker can query the target with $x_{\setminus S}$ and obtain predictive logits or token distributions at the masked positions.
We do not require access to target parameters, gradients, or internal activations.
For the main \method{} setting, the corresponding pre-fine-tuning checkpoint $\Mref$ is available under the same interface.
We additionally study a target-only setting in \cref{sec:target_only}, where only $\Mtgt$ is available.
Detailed access assumptions are provided in \cref{app:access_assumptions}.

\paragraph{Scoring and evaluation.}
When the reference model is available, we define the token-level reconstruction advantage
\begin{equation}
\label{eq:gap_def}
  \Delta_i(S)
  =
  \log p_{\Mtgt}(x_i \mid x_{\setminus S})
  -
  \log p_{\Mref}(x_i \mid x_{\setminus S}),
  \qquad i\in S.
\end{equation}
All $|S|$ gaps are obtained with one target and one reference forward pass.
Target-only attacks instead construct membership scores from the target reconstruction outputs alone.
In both settings, the resulting sample-level score is thresholded for membership prediction.
We report ROC-AUC, TPR at FPR $\in\{10\%,1\%,0.1\%\}$, and model forward evaluations per sample.

\section{MIA on Fine-Tuned
\texorpdfstring{\NoCaseChange{\textnormal{d}LLMs}}{dLLMs}
with a Reference Model}
\label{sec:method}

\begin{figure*}[t]
    \centering
    \includegraphics[width=\textwidth]{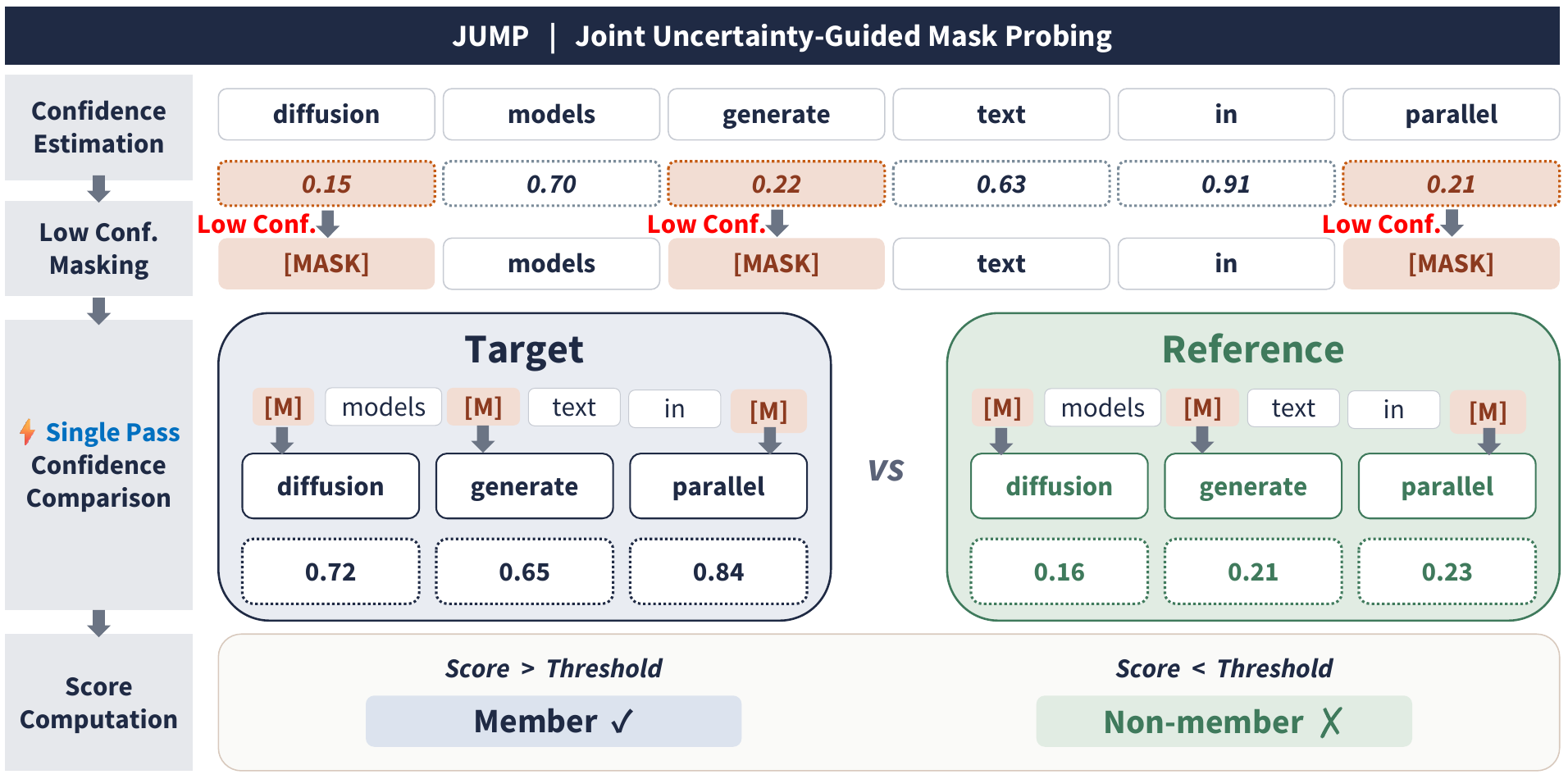}
    \caption{
    Overview of \method{}.
    The attack first selects informative positions under the reference model, then masks the selected positions jointly and scores them with the target and reference dLLMs.
    The final MIA statistic is computed from clipped token-level target/reference reconstruction gaps.
    In the visualization, general tokens are indicated by solid-line boxes, while confidence scores are denoted by dashed-line boxes.
    }
    \label{fig:pipeline}
\end{figure*}

\method{} has two variants depending on the available model access.
In this section, we consider membership inference for fine-tuned dLLMs when a reference model is available.
Specifically, we use the model before fine-tuning as the reference model.
We first identify where membership signal appears (\cref{subsec:motiv}), then define the JUMP scoring rule (\cref{subsec:method_def}) and efficient token selector (\cref{subsec:prism}).
We consider the target-only setting without a reference model separately in \cref{sec:target_only}.

\subsection{Low-confidence tokens and membership signal}
\label{subsec:motiv}

The any-order interface in dLLM raises a basic design question:
which token positions should an attacker probe?
We first study the ideal one-hole reference low-confidence score
\begin{equation}
\label{eq:ideal_q}
  q^{\star}(i\mid x)
  = \log p_{\Mref}(x_i\mid x_{\setminus\{i\}}),
\end{equation}
where smaller values indicate lower reference confidence in the true token under bidirectional context.
This quantity is useful for analysis because it directly measures how uncertain the reference dLLM is about the true token $x_i$ when only $x_i$ is masked.

\Needspace{18\baselineskip}
\begin{wrapfigure}[16]{r}{0.46\textwidth}
  \vspace{-8pt}
  \centering
  \includegraphics[width=0.78\linewidth]{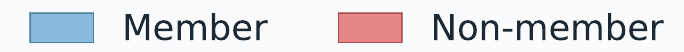}\\[0.35em]
  \includegraphics[width=\linewidth]{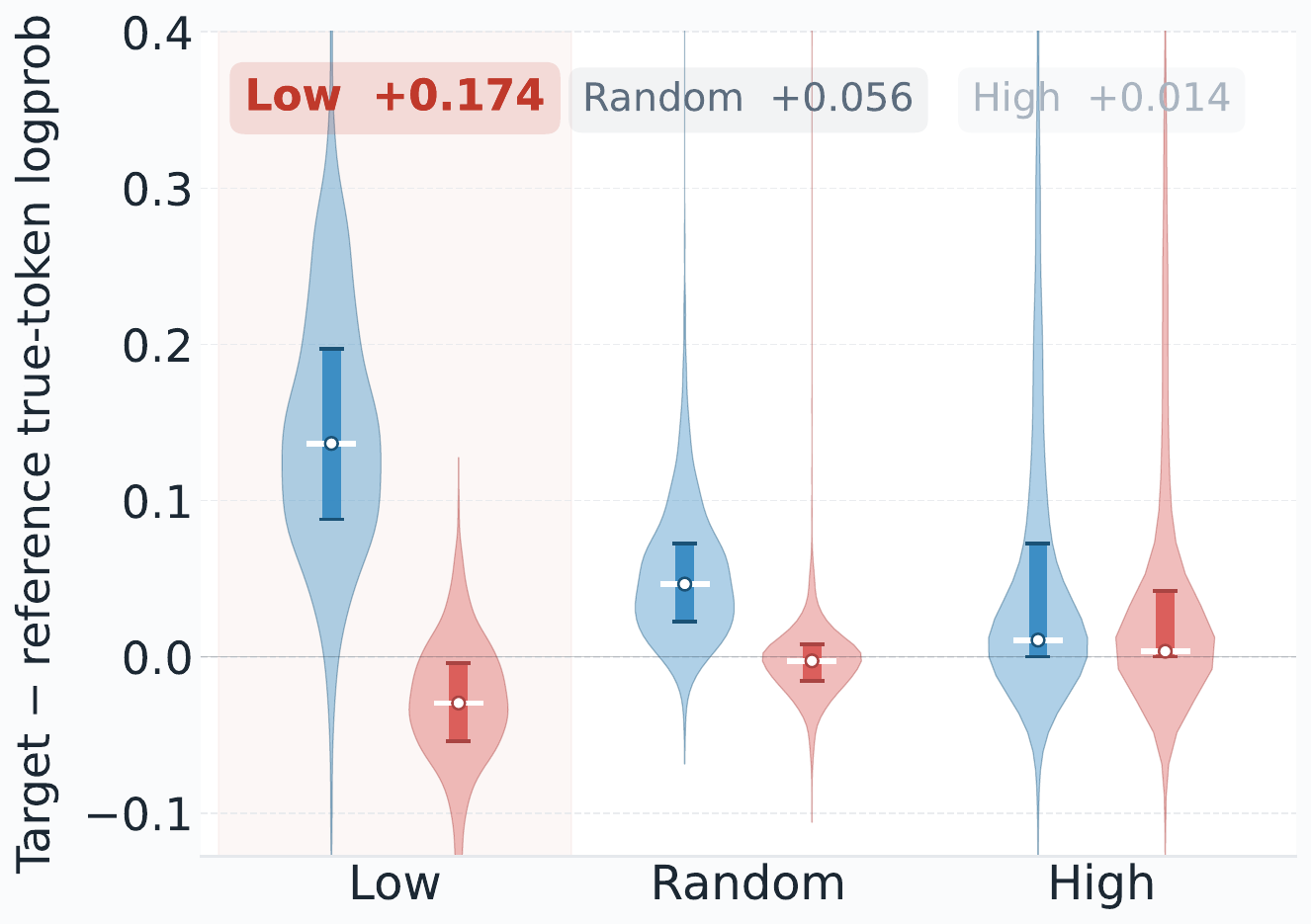}
  \caption{One-hole reconstruction gap across reference-confidence tiers.
  Low-confidence positions show the strongest separation.}
  \label{fig:motiv}
  \vspace{-7pt}
\end{wrapfigure}

To test whether this low-confidence score is related to membership signal, we run a diagnostic one-hole experiment on LLaDA-8B-Base fine-tuned on the Wikipedia (en) domain. For each candidate sequence, we use a 512-token evaluation window and, for every valid position $i$, mask only that position, i.e., $S=\{i\}$. We then record the one-hole reference probability $p_{\Mref}(x_i| x_{\setminus\{i\}})$ and the target/reference gap $\Delta_i(\{i\})$ from \cref{eq:gap_def}. Positions are ranked by \cref{eq:ideal_q} and partitioned into three groups of size $K=64$: lowest-confidence positions, uniformly random positions, and highest-confidence positions.

\cref{fig:motiv} shows that low-confidence positions carry substantially stronger membership signal.
Under one-hole masking, the fine-tuned target reconstructs these positions much better on member sequences than on non-member sequences, producing a mean member/non-member gap difference of $+0.174$.
Random and high-confidence positions show much weaker separation.
This supports the use of reference-model confidence as a localization prior: positions where the reference model has low confidence are more informative for fine-tuning membership. Low reference confidence is not a per-token guarantee: in a six-domain audit, 23.7\% of selected tokens have $|\Delta_i|\leq0.05$, yet the aggregated gap between members and non-members remains positive in every domain (\cref{app:low_confidence_gap}).

The diagnostic procedure itself is not an efficient attack.
Ranking all positions by \cref{eq:ideal_q} requires $L$ one-hole queries to the reference model.
Reusing those reference outputs, computing the corresponding gaps requires another $L$ one-hole queries to the target, for $2L$ model forwards in total.
This motivates an efficient attack that first identifies low-confidence positions and then scores them together through a single joint mask.

\subsection{\methodfull{}}
\label{subsec:method_def}

We propose \methodfull{}, which turns the low-confidence localization principle above into an efficient membership inference attack through uncertainty-guided position selection and joint masked scoring.
Given a sequence $x$ of length $L$, let $[L]\subseteq\{1,\ldots,L\}$ denote the valid positions that may be masked and scored.
\method{} uses a selector score $q_\phi(i\mid x)$ for each $i\in[L]$.
The score $q_\phi(i\mid x)$ is a scalar prediction of reference-model confidence in the true token at position $i$, rather than a vocabulary distribution.
Ideally, $q_\phi(i\mid x)$ approximates the one-hole confidence score $q^{\star}(i\mid x)$ in \cref{eq:ideal_q}, so smaller values indicate lower reference confidence.

For a probing budget $K$, JUMP selects the $K$ positions of lowest confidences:
\begin{equation}
\label{eq:hardset}
  \mathcal{H}_K(x;\phi)
  =
  \left\{ 
  i \in [L] \mid q_\phi(i \mid x) \text{ is one of the lowest } K \text{ values among all } i \in  [L]
  \right\}
\end{equation}

The selected positions are masked jointly, $S=\mathcal{H}_K(x;\phi)$, and the attacker computes $\Delta_i(S)$ for all $i\in S$ using \cref{eq:gap_def}.
Since all positions in $S$ are reconstructed in parallel, this joint probe returns $K$ token-level membership signals with one target and one reference query.
Thus, $K$ affects only the number of averaged token gaps, not the target/reference scoring NFE.

The final statistic clips and averages the selected token gaps:
\begin{equation}
\label{eq:score}
  g_{\methodmath}(x;\Mtgt,\Mref)
  = \frac{1}{K}\sum_{i\in\mathcal{H}_K(x;\phi)}
    \mathrm{clip}\!\left(\Delta_i(\mathcal{H}_K(x;\phi)), -\tau, \tau\right),
  \qquad \tau=\log 1.5.
\end{equation}
We clip to bound the effect of any single token, limiting heavy-tailed token gaps that can otherwise dominate the mean and create false positives at low FPR.
The threshold $\tau=\log 1.5$ caps each token's contribution at a $1.5{:}1$ probability ratio and is fixed throughout our evaluation; \cref{app:clip_operation_ablation} compares this readout with the unclipped mean.

\paragraph{Remark (NFE and efficiency).}
The joint scoring stage uses exactly two forward passes per sample, one through the target and one through the reference, independent of $K$ and sequence length.
The learned PRISM selector adds one reference-model pass, so the default attack uses three model forwards in total.
SAMA with $T$ masking steps instead uses $2T$ target/reference forwards.

\subsection{Efficient Token Selection}
\label{subsec:prism}

The remaining question is how to obtain $q_\phi$ without running the expensive one-hole diagnostic in \cref{subsec:motiv}.
In our default learned-selector variant, we use PRISM~\citep{kim2025fine} as an efficient low-confidence token selector.
Given an unmasked sequence, PRISM predicts a position-wise low-confidence score from the reference-model representations:
\begin{equation*}
  q_\phi(i\mid x)
  \approx
  q^{\star}(i\mid x)
  =
  \log p_{\Mref}(x_i\mid x_{\setminus\{i\}}).
\end{equation*}

PRISM is a lightweight per-token quality head trained on the reference dLLM using the self-correction construction of~\citet{kim2025fine}.
Given a partially masked sequence, the reference predicts the hidden tokens.
Some masks are filled with detached argmax predictions, and the quality head classifies each update as correct or incorrect.
It therefore approximates context-conditioned update correctness, a natural proxy for one-hole reconstruction confidence.
We train it with a correctness BCE and a small mask-ratio-weighted reconstruction loss.

This procedure generates token-level supervision from generic reference-side text, without membership labels, target information, or private examples.
At attack time, one reference pass returns both the unmasking probabilities and PRISM scores $q_\phi(i\mid x)$.
\method{} then selects the positions in \cref{eq:hardset} and jointly scores them as defined in \cref{eq:score}.

Because the frozen reference model generates its own supervision, PRISM does not require target-domain data or a particular auxiliary corpus.
Its training is a lightweight one-time procedure: in our setup, the selector checkpoint is obtained in under 35 minutes on four NVIDIA L40S GPUs and can then be reused across membership audits.
Training details and controls over the auxiliary corpus and training-pool size are provided in \cref{app:prism_training_details,app:selector_corpus_transfer,app:c4_sample_size_ablation}.

We also consider PRISM-Free, which removes auxiliary selector training entirely.
Instead of predicting one-hole confidence with a learned quality head, it ranks positions using the reference model's clean-input true-token probabilities, selects the lowest-confidence positions, and applies the same joint masked scoring rule.
Despite this simpler selection mechanism, PRISM-Free remains competitive with the learned selector; we report the comparison in \cref{subsec:prism_free_results}.

\section{Experiments}
\label{sec:exp}
\label{sec:experiments}

\begin{table}[!t]
  \centering
  \scriptsize
  \setlength{\tabcolsep}{2.4pt}
  \renewcommand{\arraystretch}{1.03}

  \caption{
    Per-domain reference-assisted results on LLaDA-8B-Base and Dream-v0-Base-7B.
    We report ROC-AUC and TPR at fixed FPR ($\uparrow$) with total NFE ($\downarrow$).
    SAMA uses $T=16$ (32 forwards) and \method{} uses 3; best quality values are in \textbf{bold}.
  }
  \label{tab:main_results}

  \begin{tabular}{lccccc ccccc ccccc}
  \toprule

  \multicolumn{16}{c}{\small\textbf{LLaDA-8B-Base}} \\
  \midrule

  & \multicolumn{5}{c}{\textbf{ArXiv}}
  & \multicolumn{5}{c}{\textbf{GitHub}}
  & \multicolumn{5}{c}{\textbf{HackerNews}} \\
  \cmidrule(lr){2-6}\cmidrule(lr){7-11}\cmidrule(lr){12-16}

  \textbf{Method}
  & \multicolumn{4}{c}{Attack quality ($\uparrow$)} & Cost ($\downarrow$)
  & \multicolumn{4}{c}{Attack quality ($\uparrow$)} & Cost ($\downarrow$)
  & \multicolumn{4}{c}{Attack quality ($\uparrow$)} & Cost ($\downarrow$) \\
  \cmidrule(lr){2-5}\cmidrule(lr){6-6}
  \cmidrule(lr){7-10}\cmidrule(lr){11-11}
  \cmidrule(lr){12-15}\cmidrule(lr){16-16}

  & AUC & T@10 & T@1 & T@0.1 & NFE
  & AUC & T@10 & T@1 & T@0.1 & NFE
  & AUC & T@10 & T@1 & T@0.1 & NFE \\
  \midrule

  Loss
  & 0.52 & 0.12 & 0.01 & 0.00 & 16
  & 0.59 & 0.20 & 0.04 & 0.01 & 16
  & 0.51 & 0.11 & 0.01 & 0.01 & 16 \\

  ZLIB
  & 0.52 & 0.13 & 0.01 & 0.00 & 16
  & 0.61 & 0.23 & 0.07 & 0.02 & 16
  & 0.51 & 0.10 & 0.01 & 0.01 & 16 \\

  SAMA
  & 0.82 & 0.54 & 0.28 & 0.07 & 32
  & 0.77 & 0.43 & 0.14 & 0.07 & 32
  & 0.71 & 0.34 & 0.05 & 0.01 & 32 \\

  \rowcolor{oursrow}
  \textbf{\method{}}
  & \textbf{0.94} & \textbf{0.84} & \textbf{0.55} & \textbf{0.30} & 3
  & \textbf{0.86} & \textbf{0.66} & \textbf{0.32} & \textbf{0.18} & 3
  & \textbf{0.76} & \textbf{0.39} & \textbf{0.10} & \textbf{0.02} & 3 \\

  \midrule[\heavyrulewidth]

  & \multicolumn{5}{c}{\textbf{PubMed Central}}
  & \multicolumn{5}{c}{\textbf{Wikipedia (en)}}
  & \multicolumn{5}{c}{\textbf{Pile-CC}} \\
  \cmidrule(lr){2-6}\cmidrule(lr){7-11}\cmidrule(lr){12-16}

  \textbf{Method}
  & \multicolumn{4}{c}{Attack quality ($\uparrow$)} & Cost ($\downarrow$)
  & \multicolumn{4}{c}{Attack quality ($\uparrow$)} & Cost ($\downarrow$)
  & \multicolumn{4}{c}{Attack quality ($\uparrow$)} & Cost ($\downarrow$) \\
  \cmidrule(lr){2-5}\cmidrule(lr){6-6}
  \cmidrule(lr){7-10}\cmidrule(lr){11-11}
  \cmidrule(lr){12-15}\cmidrule(lr){16-16}

  & AUC & T@10 & T@1 & T@0.1 & NFE
  & AUC & T@10 & T@1 & T@0.1 & NFE
  & AUC & T@10 & T@1 & T@0.1 & NFE \\
  \midrule

  Loss
  & 0.52 & 0.13 & 0.01 & 0.00 & 16
  & 0.51 & 0.10 & 0.01 & 0.00 & 16
  & 0.51 & 0.11 & 0.01 & 0.00 & 16 \\

  ZLIB
  & 0.52 & 0.13 & 0.01 & 0.00 & 16
  & 0.51 & 0.10 & 0.01 & 0.00 & 16
  & 0.52 & 0.13 & 0.02 & 0.00 & 16 \\

  SAMA
  & 0.81 & 0.51 & 0.23 & 0.03 & 32
  & 0.91 & 0.73 & 0.47 & 0.36 & 32
  & 0.90 & 0.70 & 0.40 & \textbf{0.20} & 32 \\

  \rowcolor{oursrow}
  \textbf{\method{}}
  & \textbf{0.92} & \textbf{0.75} & \textbf{0.43} & \textbf{0.12} & 3
  & \textbf{0.98} & \textbf{0.95} & \textbf{0.80} & \textbf{0.67} & 3
  & \textbf{0.96} & \textbf{0.90} & \textbf{0.60} & 0.10 & 3 \\

  \midrule[\heavyrulewidth]
  \multicolumn{16}{c}{\small\textbf{Dream-v0-Base-7B}} \\
  \midrule

  & \multicolumn{5}{c}{\textbf{ArXiv}}
  & \multicolumn{5}{c}{\textbf{GitHub}}
  & \multicolumn{5}{c}{\textbf{HackerNews}} \\
  \cmidrule(lr){2-6}\cmidrule(lr){7-11}\cmidrule(lr){12-16}

  \textbf{Method}
  & \multicolumn{4}{c}{Attack quality ($\uparrow$)} & Cost ($\downarrow$)
  & \multicolumn{4}{c}{Attack quality ($\uparrow$)} & Cost ($\downarrow$)
  & \multicolumn{4}{c}{Attack quality ($\uparrow$)} & Cost ($\downarrow$) \\
  \cmidrule(lr){2-5}\cmidrule(lr){6-6}
  \cmidrule(lr){7-10}\cmidrule(lr){11-11}
  \cmidrule(lr){12-15}\cmidrule(lr){16-16}

  & AUC & T@10 & T@1 & T@0.1 & NFE
  & AUC & T@10 & T@1 & T@0.1 & NFE
  & AUC & T@10 & T@1 & T@0.1 & NFE \\
  \midrule

  Loss
  & 0.54 & 0.13 & 0.01 & 0.00 & 16
  & 0.60 & 0.21 & 0.05 & 0.01 & 16
  & 0.54 & 0.12 & 0.01 & 0.01 & 16 \\

  ZLIB
  & 0.54 & 0.14 & 0.01 & 0.00 & 16
  & 0.63 & 0.24 & 0.07 & 0.03 & 16
  & 0.54 & 0.11 & 0.02 & 0.01 & 16 \\

  SAMA
  & 0.84 & 0.57 & 0.29 & 0.22 & 32
  & 0.83 & 0.62 & 0.26 & \textbf{0.12} & 32
  & 0.83 & 0.57 & 0.17 & 0.05 & 32 \\

  \rowcolor{oursrow}
  \textbf{\method{}}
  & \textbf{0.93} & \textbf{0.81} & \textbf{0.52} & \textbf{0.32} & 3
  & \textbf{0.91} & \textbf{0.75} & \textbf{0.36} & 0.01 & 3
  & \textbf{0.93} & \textbf{0.81} & \textbf{0.49} & \textbf{0.23} & 3 \\

  \midrule[\heavyrulewidth]

  & \multicolumn{5}{c}{\textbf{PubMed Central}}
  & \multicolumn{5}{c}{\textbf{Wikipedia (en)}}
  & \multicolumn{5}{c}{\textbf{Pile-CC}} \\
  \cmidrule(lr){2-6}\cmidrule(lr){7-11}\cmidrule(lr){12-16}

  \textbf{Method}
  & \multicolumn{4}{c}{Attack quality ($\uparrow$)} & Cost ($\downarrow$)
  & \multicolumn{4}{c}{Attack quality ($\uparrow$)} & Cost ($\downarrow$)
  & \multicolumn{4}{c}{Attack quality ($\uparrow$)} & Cost ($\downarrow$) \\
  \cmidrule(lr){2-5}\cmidrule(lr){6-6}
  \cmidrule(lr){7-10}\cmidrule(lr){11-11}
  \cmidrule(lr){12-15}\cmidrule(lr){16-16}

  & AUC & T@10 & T@1 & T@0.1 & NFE
  & AUC & T@10 & T@1 & T@0.1 & NFE
  & AUC & T@10 & T@1 & T@0.1 & NFE \\
  \midrule

  Loss
  & 0.54 & 0.14 & 0.03 & 0.00 & 16
  & 0.53 & 0.12 & 0.01 & 0.01 & 16
  & 0.53 & 0.13 & 0.01 & 0.00 & 16 \\

  ZLIB
  & 0.53 & 0.14 & 0.03 & 0.00 & 16
  & 0.53 & 0.13 & 0.02 & 0.01 & 16
  & 0.54 & 0.14 & 0.02 & 0.00 & 16 \\

  SAMA
  & 0.84 & 0.59 & 0.30 & 0.16 & 32
  & 0.86 & 0.62 & 0.35 & 0.25 & 32
  & 0.90 & 0.72 & 0.35 & 0.19 & 32 \\

  \rowcolor{oursrow}
  \textbf{\method{}}
  & \textbf{0.95} & \textbf{0.86} & \textbf{0.53} & \textbf{0.35} & 3
  & \textbf{0.96} & \textbf{0.89} & \textbf{0.63} & \textbf{0.29} & 3
  & \textbf{0.97} & \textbf{0.94} & \textbf{0.77} & \textbf{0.38} & 3 \\

  \bottomrule
  \end{tabular}
\end{table}

\paragraph{Setup.}
We evaluate LLaDA-8B-Base~\citep{nie2025large} and Dream-v0-Base-7B~\citep{ye2025dream} on six MIMIR domains: ArXiv, GitHub, HackerNews, Pile-CC, PubMed Central, and Wikipedia (en)~\citep{duan2024membership}.
For each model and domain, the target is fine-tuned on 1{,}000 MIMIR training documents (members) and evaluated against 1{,}000 documents from the corresponding MIMIR test split (non-members).
Reference-assisted attacks use the exact pre-fine-tuning checkpoint, and all attacks share the same target checkpoint and candidates.

\paragraph{Metrics and configurations.}
We compare with Loss, ZLIB~\citep{carlini2021extracting}, and SAMA~\citep{chen2026membership}, and report ROC-AUC, TPR at FPR $\in\{10\%,1\%,0.1\%\}$, and the total number of model forward evaluations (NFE).
\method{} uses a C4-trained PRISM selector, $K=64$, and $\tau=\log 1.5$.
Metrics are computed independently within each domain and then averaged with equal weight; samples are never pooled across domains or model families.
Model-specific training and checkpoint details appear in \cref{app:target_training_setup}, and attack and metric details in \cref{app:baseline_configurations,app:metric_protocol}.

\subsection{Main Results}
\label{subsec:performance}

\Cref{tab:main_results} shows that \method{} improves AUC over SAMA in every domain.
Mean AUC rises from $0.8193$ to $0.9024$ on LLaDA and from $0.8513$ to $0.9415$ on Dream.
On LLaDA, mean TPR@1\% FPR increases from $0.2625$ to $0.4663$ and TPR@0.1\% FPR from $0.1220$ to $0.2320$; Dream shows the same aggregate improvement. Domain-level estimates at 0.1\% FPR should be interpreted cautiously:
with 1{,}000 non-members per domain, this operating point is determined
by roughly one false positive and therefore has coarse tail resolution;
the corresponding bootstrap intervals are reported in
\cref{app:bootstrap_ci}.

The table also reports evaluation cost directly.
SAMA with $T=16$ uses 16 target and 16 reference evaluations, for 32 NFE per document.
\method{} uses one reference pass for selection and one target/reference scoring pair, for 3 NFE.
The selected-token budget $K$ changes the number of token gaps returned by the joint query, not the scoring NFE.
Measured runtime falls from $1.863$ seconds per sample for SAMA to $0.184$ for \method{}.

The trend persists in lower-exposure and mixed-domain controls: \method{} reaches $0.7316$ versus SAMA's $0.6600$ at two epochs and $0.7466$ versus $0.7116$ on the mixed-domain target.
Bootstrap intervals and selector, exposure, and clipping controls are reported in \cref{app:bootstrap_ci,app:selector_controls,app:exposure_data_controls,app:clip_and_budget}.

\subsection{PRISM-Free results}
\label{subsec:prism_free_results}

\begin{table}[H]
\centering
\small
\setlength{\tabcolsep}{5.4pt}
\renewcommand{\arraystretch}{1.05}
\caption{PRISM-Free equal-weight six-domain means.}
\label{tab:prism_free}
\begin{tabular}{llrrrr}
\toprule
Family & Method & AUC & T@10 & T@1 & T@0.1 \\
\midrule
\multirow{3}{*}{LLaDA}
& SAMA & 0.8193 & 0.5393 & 0.2625 & 0.1220 \\
& PRISM-Free & 0.8677 & 0.6527 & 0.3238 & 0.1443 \\
\rowcolor{oursrow}
& \textbf{\method{}} & \textbf{0.9024} & \textbf{0.7500} & \textbf{0.4663} & \textbf{0.2320} \\
\midrule
\multirow{3}{*}{Dream}
& SAMA & 0.8513 & 0.6122 & 0.2862 & 0.1660 \\
& PRISM-Free & 0.9404 & 0.8322 & 0.4997 & 0.2198 \\
\rowcolor{oursrow}
& \textbf{\method{}} & \textbf{0.9415} & \textbf{0.8438} & \textbf{0.5505} & \textbf{0.2653} \\
\bottomrule
\end{tabular}
\end{table}

PRISM-Free replaces the learned PRISM head with the reference model's clean-input true-token score and therefore requires no auxiliary selector training.  It improves mean ROC-AUC over SAMA on both LLaDA ($0.8677$ versus $0.8193$) and Dream ($0.9404$ versus $0.8513$).  On Dream it nearly matches \method{} in mean AUC ($0.9404$ versus $0.9415$), while the learned selector remains stronger at low FPR and provides a larger gain on LLaDA.  Thus, clean-text reference confidence is already a useful proxy for reconstruction difficulty, while the learned position-quality head makes the selected evidence more consistent.  PRISM-Free still uses the exact pre-fine-tuning model for score calibration and is therefore reference-assisted, not target-only.

\section{MIA on Fine-Tuned
\texorpdfstring{\NoCaseChange{\textnormal{d}LLMs}}{dLLMs}
without a Reference Model}
\label{sec:target_only}

We now consider membership inference when no reference model is available.
Without a reference, the target--reference reconstruction difference used by \method{} is unavailable.
Inspired by the token-level likelihood calibration of Min-K\%++~\citep{zhang2025minkpp}, we develop a target-only approach that extracts membership evidence from the target's own token predictions.
Our approach focuses on relative token-level signals, filters out overly confident predictions, and exploits the parallel decoding of dLLMs for efficient probing.

\subsection{Relative Token Preference}
\label{subsec:to_relative_preference}

We begin with an ideal one-hole probe at position $i$, where only the observed token $x_i$ is masked. Let
\begin{equation*}
  q_i
  = \log p_{\Mtgt}(x_i \mid x_{\setminus\{i\}})
\end{equation*}
denote its reconstruction log-likelihood. Without a reference model, however, $q_i$ alone is ambiguous: a low likelihood may arise either because $x_i$ is intrinsically difficult to predict or because the target model has not learned a strong preference for it.

We therefore measure the likelihood of $x_i$ relative to the model's own predictions at the same position. To capture the model's uncertainty over alternative tokens, we use the predictive entropy at position $i$:
\begin{equation*}
  H_i
  = -\sum_{v \in \mathcal{A}}
  p_{\Mtgt}(v \mid x_{\setminus\{i\}})
  \log p_{\Mtgt}(v \mid x_{\setminus\{i\}}).
\end{equation*}
Combining this term with $q_i$, we define $q_i+H_i$ as the \emph{relative token preference}. This score admits the following expected log-likelihood-ratio interpretation:
\begin{equation*}
  q_i+H_i
  =
  \mathbb{E}_{v \sim p_{\Mtgt}(\cdot \mid x_{\setminus\{i\}})}
  \left[
    \log
    \frac{p_{\Mtgt}(x_i \mid x_{\setminus\{i\}})}
         {p_{\Mtgt}(v \mid x_{\setminus\{i\}})}
  \right].
\end{equation*}
Thus, $q_i+H_i$ measures how strongly the target model favors the observed token $x_i$ over alternatives. Training on the candidate sequence strengthen the model's preference for $x_i$ over competing tokens. We use $q_i+H_i$ as a token-level membership statistic and assess its discriminative value empirically.

\subsection{Filtering Near-Deterministic Reconstructions}
\label{subsec:to_saturation}

Real text contains many near-deterministic positions where the surrounding context alone can drive the true-token probability close to one. At such positions, the model can already assign very high probability to the observed token without having trained on the candidate sequence. These near-deterministic reconstructions can therefore yield weak membership signals even for member sequences. Including many such positions can dilute useful evidence in the document-level average.

This behavior becomes clear in the near-deterministic limit:
\begin{equation*}
  p_{\Mtgt}(x_i \mid x_{\setminus\{i\}}) \rightarrow 1
  \quad\Longrightarrow\quad
  q_i \rightarrow 0,\qquad
  H_i \rightarrow 0,\qquad
  q_i+H_i \rightarrow 0.
\end{equation*}
When the predictive mass is already concentrated on $x_i$, both the reconstruction log-likelihood and predictive entropy saturate, causing the relative token preference to vanish.

We therefore exclude positions whose one-hole true-token probability is larger than $0.99$. This filtering does not select the hardest or lowest-confidence tokens; it only removes positions where the reconstruction has effectively saturated. We use the same threshold across all domains and both model families. Reconstruction-confidence distributions, saturation diagnostics, and threshold sensitivity are reported in \cref{app:reference_free_ablations}.

\subsection{Efficient Grouped Approximation}
\label{subsec:to_grouped}

The target-only formulation is defined from one-hole reconstruction quantities, so we would ideally evaluate $q_i$ and $H_i$ with only $x_i$ masked. Exact one-hole evaluation, however, requires a separate masked query for every valid token position and is therefore expensive for practical auditing. One might instead use PRISM~\citep{kim2025fine}, as in the reference-assisted setting, to approximate token-level reconstruction difficulty. However, PRISM relies on model representations: in our main attack these are obtained from the reference model, which is unavailable here, while applying PRISM to the target would require internal target access beyond our predictive-output-only threat model. We therefore approximate the one-hole quantities directly with grouped masked queries.

Inspired by the cyclic sampling strategy of Q-Skew~\citep{zhai2026membership}, we partition the valid positions into 16 disjoint, approximately equal-sized groups $S_1,\ldots,S_{16}$ and mask each group once. For every $i\in S_g$, the corresponding target query provides
\begin{align*}
  q_i(S_g)
  &=
  \log p_{\Mtgt}(x_i\mid x_{\setminus S_g}),\\
  H_i(S_g)
  &=
  -\sum_{v\in\mathcal{A}}
  p_{\Mtgt}(v\mid x_{\setminus S_g})
  \log p_{\Mtgt}(v\mid x_{\setminus S_g}).
\end{align*}
These grouped quantities approximate the one-hole $q_i$ and $H_i$, with the other positions in $S_g$ masked simultaneously. Because all masked positions are reconstructed in parallel, all valid positions are covered with only 16 target queries.

Let $g(i)$ denote the unique group index such that $i\in S_{g(i)}$. We apply the filtering rule from \cref{subsec:to_saturation} using the grouped reconstruction probabilities and denote the retained positions by $\mathcal{R}(x)$. Target-Only \method{} then scores a document by averaging their relative token preferences:
\begin{equation*}
  g_{\mathrm{TO\text{-}JUMP}}(x;\Mtgt)
  =
  \frac{1}{|\mathcal{R}(x)|}
  \sum_{i\in\mathcal{R}(x)}
  \left[
    q_i(S_{g(i)})+H_i(S_{g(i)})
  \right].
\end{equation*}
Higher values indicate stronger membership evidence. Since filtering and scoring use the same grouped target outputs, no additional model evaluations are required.

\subsection{Target-Only Evaluation}
\label{subsec:target_only_eval}

\begin{table}[t]
\centering
\small
\setlength{\tabcolsep}{5.8pt}
\renewcommand{\arraystretch}{1.03}

\caption{
Target-only results (ROC-AUC / TPR@10\% FPR, $\uparrow$).
Random uses uniformly sampled positions;
Clean Score selects positions by clean-input target true-token log-probability,
while Ref.\ Guided uses reference-selected positions as a control.
Masked Score and Target-Only \method{} share the same 16-way grouped outputs
and $p_{\mathrm{true}}\leq 0.99$ filtering rule;
Masked Score uses the true-token log-probability $q_i$,
while Target-Only \method{} uses the relative token preference $q_i+H_i$.
Means are macro-averaged across domains.
Best and second-best are bold and underlined.
}
\label{tab:reference_free_results}

\makebox[\linewidth][c]{%
\begin{tabular}{l*{7}{c}}
\toprule

\multicolumn{8}{c}{\textbf{LLaDA-8B-Base}} \\
\midrule
\textbf{Method}
& \textbf{ArXiv}
& \textbf{GitHub}
& \textbf{HackerNews}
& \textbf{PubMed}
& \textbf{Wikipedia}
& \textbf{Pile-CC}
& \textbf{Mean} \\
\midrule

Random
& .558/.112
& \underline{.584}/.158
& .534/.120
& .553/.142
& .553/.124
& .537/.108
& .553/.127 \\

Clean Score
& .574/.128
& .564/.174
& .530/.121
& .559/.131
& .580/.134
& .555/.142
& .560/.138 \\

Ref. Guided
& .576/.172
& .578/\underline{.179}
& .533/\underline{.127}
& .578/.136
& .559/.137
& .546/.126
& .562/.146 \\

Masked Score
& \underline{.602}/\underline{.176}
& .577/.175
& \underline{.541}/.116
& \underline{.598}/\underline{.186}
& \underline{.604}/\underline{.171}
& \underline{.591}/\underline{.176}
& \underline{.585}/\underline{.167} \\

\rowcolor{oursrow}
\textbf{TO-\method{}}
& \textbf{.617/.214}
& \textbf{.596/.206}
& \textbf{.566/.148}
& \textbf{.621/.211}
& \textbf{.638/.221}
& \textbf{.619/.197}
& \textbf{.609/.200} \\

\midrule

\multicolumn{8}{c}{\textbf{Dream-v0-Base-7B}} \\
\midrule
\textbf{Method}
& \textbf{ArXiv}
& \textbf{GitHub}
& \textbf{HackerNews}
& \textbf{PubMed}
& \textbf{Wikipedia}
& \textbf{Pile-CC}
& \textbf{Mean} \\
\midrule

Random
& .600/.180
& .632/.183
& .581/.169
& .571/.139
& .587/.160
& .568/.158
& .590/.165 \\

Clean Score
& .631/.191
& \textbf{.638/.238}
& .611/.207
& .600/.164
& .616/\underline{.210}
& .591/.184
& .614/.199 \\

Ref. Guided
& .630/.204
& .636/.233
& .609/.186
& .616/\underline{.209}
& .606/.193
& .610/.194
& .618/\underline{.203} \\

Masked Score
& \underline{.658}/\textbf{.239}
& \underline{.637}/\underline{.237}
& \underline{.653}/\underline{.217}
& \underline{.633}/\textbf{.222}
& \underline{.647}/\textbf{.251}
& \textbf{.644}/\underline{.224}
& \textbf{.645/.232} \\

\rowcolor{oursrow}
\textbf{TO-\method{}}
& \textbf{.659}/\underline{.228}
& .581/.129
& \textbf{.657/.219}
& \textbf{.636}/.200
& \textbf{.653}/.194
& \underline{.639}/\textbf{.228}
& \underline{.638}/.200 \\

\bottomrule
\end{tabular}%
}
\end{table}

We reuse the target checkpoints and the per-domain sets of 1,000 members
and 1,000 non-members from \cref{sec:experiments}, but do not use the
pre-fine-tuning checkpoint.
Metrics are computed independently within each domain and then macro-averaged.

\Cref{tab:reference_free_results} reports target-only membership inference
performance across the six MIMIR domains on LLaDA and Dream.
The first three rows compare position-selection strategies.
Random positions and clean-confidence selection use only the final target
model, while reference-guided selection uses a reference-derived position set
but removes target--reference score calibration and is included only as a
calibration control.
The final two rows use the same 16-way grouped outputs and
$p_{\mathrm{true}}\leq 0.99$ filtering rule, allowing us to compare raw masked
log-probability with relative token preference.

Target-Only \method{} achieves mean ROC-AUCs of $0.6095$ on LLaDA and
$0.6375$ on Dream.
It outperforms the random and clean-confidence target-only baselines on both
model families, showing that membership information can be extracted from the
target model's masked reconstruction distribution even without a
pre-fine-tuning reference.
On LLaDA, relative token preference improves over masked log-probability in
all six domains, increasing the mean AUC from $0.5854$ to $0.6095$.

On Dream, masked log-probability is stronger on average
($0.6454$ versus $0.6375$), with the largest difference on GitHub.
Nevertheless, Target-Only \method{} remains stronger than the target-only
position-selection baselines and improves over masked log-probability in four
of the other five domains.
Additional analyses of reconstruction confidence, filtering thresholds, and
score normalization are provided in \cref{app:reference_free_ablations}.

\section{Conclusion}
\label{sec:conclusion}

\method{} turns the dLLM interface into a membership-auditing framework by identifying informative positions and evaluating them through parallel reconstruction.
When the pre-fine-tuning checkpoint is available, \method{} localizes low-confidence positions and measures their fine-tuning-induced reconstruction gains against the reference, improving mean AUC to $0.902$ on LLaDA and $0.942$ on Dream while reducing SAMA's 32 NFE to 3.
When only the final target is available, Target-Only \method{} uses grouped masking, filters near-deterministic positions, and aggregates relative token preference, reaching mean AUCs of $0.609$ and $0.638$.
Together, these results show that dLLM reconstruction signals support membership auditing both with and without access to a reference.

\subsection*{AI use statement}
Generative AI tools were used only to improve the writing, formatting, and presentation of the manuscript.
The authors reviewed all AI-assisted edits and take full responsibility for the final content of the paper.

\subsection*{Ethics statement}
Membership inference is dual use: it can support authorized privacy auditing, but it can also reveal whether a record participated in model adaptation.
We evaluate publicly available model families on established benchmark corpora, report aggregate results, and do not release inferred membership decisions for identifiable records.
The intended use of this work is controlled auditing by model developers, data owners, and authorized evaluators.

\subsection*{Reproducibility statement}
The main text defines the attack scores, mask construction, and query accounting.
The appendix specifies the data splits, target-training protocol, model-access assumptions, hyperparameters, baselines, metrics, bootstrap procedure, runtime, ablations, and asset licenses.
Domain-wise results are reported separately for both model families so that the stated averages can be reconstructed directly.

\bibliographystyle{iclr2027_conference}
\bibliography{references}

\clearpage
\appendix
\pretocmd{\subsection}{\Needspace{4.5\baselineskip}}{}{}

\section{Experimental Details}
\label[appendix]{app:experimental_details}
\subsection{Threat Model and Access Assumptions}
\label[appendix]{app:access_assumptions}

We consider two fine-tuning membership-auditing settings.
In the primary reference-assisted setting, the corresponding pre-fine-tuning checkpoint is available, as in open-weight adaptation and developer-side privacy audits.
Target-Only JUMP covers the complementary setting in which the paired pre-fine-tuning checkpoint is unavailable.

Reference-assisted access corresponds to two concrete scenarios.
In known-lineage open-weight adaptation, a published base checkpoint is used to produce a disclosed derivative model.
In authorized developer-side auditing, the model owner retains both the base and adapted checkpoints.

The default learned PRISM selector additionally uses reference-model representations during one-time offline training on generic public text; it uses no target parameters or gradients, private fine-tuning examples, membership labels, or evaluation examples.
PRISM-Free removes this auxiliary training and hidden-state access and uses only model output scores for position selection and final scoring.
It still uses the pre-fine-tuning reference and is therefore reference-assisted rather than target-only.

\subsection{Original MIMIR Protocol and Member Definition}
\label[appendix]{app:mimir_protocol}

We use ArXiv, GitHub, HackerNews, Pile-CC, PubMed Central, and Wikipedia (en) from MIMIR.
For each domain, the 1,000 fine-tuning documents constitute the member set, and the corresponding 1,000 MIMIR test documents constitute the non-member set.
We verify row order and content hashes against the target train/test manifests and find zero exact member/non-member hash overlap in all six domains.

\subsection{Target Model Training and Checkpoint Selection}
\label{app:target_training_setup}

\paragraph{LLaDA.}
Each LLaDA target is initialized from \texttt{GSAI-ML/LLaDA-8B-Base} and fine-tuned separately on one domain's 1{,}000 member documents.
We use AdamW with weight decay $0.1$ and a configured learning rate of $5\times10^{-5}$.
The scheduler is linear with 500 optimizer-update warmup steps.
Because the training runs contain at most 80 optimizer updates, the realized schedule remains within the warmup phase and reaches a learning rate of $8\times10^{-6}$ at step 80.
We use maximum sequence length 512 and random seed 42.

Training uses four GPUs in bf16 with DeepSpeed ZeRO-3 and gradient checkpointing.
The per-device batch size is 1 with 12 gradient-accumulation steps, giving an
effective batch size of 48.
All LLaDA targets are trained with the same maximum four-epoch budget.
We evaluate checkpoints at approximately one-epoch intervals and select the
checkpoint with the lowest validation loss.
Here the validation set is the corresponding MIMIR test split, which also serves
as the non-member evaluation set.
These examples receive no gradient updates, but they can influence which
checkpoint is retained.
This selects checkpoint 61 for GitHub and HackerNews and checkpoint 80 for
ArXiv, Pile-CC, PubMed Central, and Wikipedia, corresponding to approximately
three and four epochs, respectively.
Checkpoint suffixes denote optimizer global steps.
We report the validation-set reuse explicitly because the selected checkpoints
are used for the main results.
The fixed epoch-1 and epoch-2 controls disable evaluation-based checkpoint
selection and use the predetermined endpoints at steps 20 and 40.

\paragraph{Dream.}
Dream targets are initialized from \texttt{Dream-org/Dream-v0-Base-7B} and use the same data size, optimizer, learning-rate schedule, sequence length, seed, distributed training setup, batch size, and gradient accumulation as LLaDA.
All six Dream domains use checkpoint 80, corresponding to approximately four epochs.

\paragraph{Checkpoint-selection robustness.}
GitHub and HackerNews are the only two LLaDA domains for which the common
checkpoint-selection rule retains an earlier checkpoint. To test whether the
reported results depend on these earlier endpoints, we continue the same
training trajectories to the predetermined checkpoint-80 endpoint and
reevaluate the attacks. \Cref{tab:checkpoint_selection_control} compares the
selected checkpoints used in the main evaluation with the fixed four-epoch
endpoints.

\begin{table*}[!htbp]
\centering
\scriptsize
\setlength{\tabcolsep}{4pt}
\renewcommand{\arraystretch}{1.05}

\caption{
Checkpoint-selection robustness on GitHub and HackerNews.
``Selected'' denotes the checkpoint used in the main evaluation
($\sim$3 epochs), while ``Epoch 4'' evaluates checkpoint 80.
}
\label{tab:checkpoint_selection_control}

\begin{tabular}{llcccccccccccc}
\toprule
& &
\multicolumn{4}{c}{SAMA} &
\multicolumn{4}{c}{JUMP} &
\multicolumn{4}{c}{Target-Only JUMP} \\
\cmidrule(lr){3-6}
\cmidrule(lr){7-10}
\cmidrule(lr){11-14}

Domain & Checkpoint
& AUC & T@10 & T@1 & T@0.1
& AUC & T@10 & T@1 & T@0.1
& AUC & T@10 & T@1 & T@0.1 \\
\midrule

\multirow{2}{*}{GitHub}
& Selected
& 0.77 & 0.43 & 0.15 & 0.07
& 0.86 & 0.66 & 0.32 & 0.18
& 0.60 & 0.21 & 0.03 & 0.01 \\

& Epoch 4
& 0.83 & 0.58 & 0.25 & 0.08
& 0.91 & 0.77 & 0.50 & 0.32
& 0.63 & 0.24 & 0.04 & 0.01 \\

\midrule

\multirow{2}{*}{HackerNews}
& Selected
& 0.71 & 0.34 & 0.05 & 0.01
& 0.76 & 0.39 & 0.10 & 0.02
& 0.57 & 0.15 & 0.02 & 0.00 \\

& Epoch 4
& 0.80 & 0.51 & 0.14 & 0.02
& 0.86 & 0.64 & 0.21 & 0.08
& 0.63 & 0.20 & 0.03 & 0.00 \\

\bottomrule
\end{tabular}
\end{table*}

The fixed four-epoch endpoints strengthen the attack signal rather than
weakening it. For both GitHub and HackerNews, JUMP and SAMA improve across
all reported metrics, while Target-Only JUMP also improves in ROC-AUC and
TPR@10\% FPR. The ordering of JUMP over SAMA is preserved at the fixed
four-epoch endpoint. We nevertheless retain the selected checkpoints in the
main evaluation to follow the common SAMA-matched checkpoint-selection rule
rather than replacing them post hoc with stronger endpoints.

\subsection{PRISM Selector Training}
\label[appendix]{app:prism_training_details}

The default LLaDA PRISM selector is trained on a materialized
200k-sample English C4 subset~\citep{raffel2020exploring}.
We split 190k training sequences and 10k held-out sequences; after preprocessing, 189,919 training sequences remain.
The selector is initialized from the exact base backbone and trained for one epoch on four L40S GPUs in fp16.
The per-GPU batch size is 4 with two gradient-accumulation steps, for an effective batch size of 32.
We use maximum length 256, AdamW with learning rate $10^{-4}$, zero weight decay, and seed 42.

Only the PRISM quality head and LoRA adapters are trained; the reference backbone is otherwise frozen.
This updates 29.38M of 8.04B parameters (0.37\%).
LoRA is applied to query, key, and value projections with rank 16, alpha 16, and dropout 0.1.
For each clean sequence $x_0$, a masking ratio
$t\sim\mathrm{Uniform}(10^{-3},1)$ is sampled, valid tokens are independently masked, and up to eight masked positions are filled with argmax predictions.
The quality head predicts whether each update matches the ground-truth token.
The retained objective is
\begin{equation*}
    \mathcal{L}_{\mathrm{PRISM}}
    =
    \mathcal{L}_{\mathrm{BCE}}
    +
    0.1\,\mathcal{L}_{\mathrm{CE}},
\end{equation*}
where the reconstruction cross-entropy term is weighted by the inverse masking ratio.
The fixed one-epoch checkpoint is used without validation-based selector checkpoint search.
Training takes 2,968 optimizer steps and approximately 1,613 seconds (26.9 minutes).

\subsection{SAMA Configuration and Forward Counting}
\label[appendix]{app:baseline_configurations}

The main SAMA baseline uses $T=16$ progressive masking steps. At each step,
the target and exact-base reference are evaluated once under the current masked
context. From the resulting token losses, the implementation samples 128 local
random subsets of 10 valid positions and computes the subset-level comparison
votes in post-processing. Thus, each step requires one target and one reference
forward pass, giving $2T=32$ NFE per example. The closest-budget control uses
$T=2$, or four forward calls. All methods use the same 512-token windows and
evaluation manifests. Wall-clock time is normalized to seconds per evaluation
sample and reported separately in Appendix D.

\subsection{Default Configuration and Metric Estimation}
\label[appendix]{app:metric_protocol}

Reference-assisted JUMP selects $K=64$ positions and applies symmetric token-wise clipping with $\tau=\log 1.5$.
Target-Only JUMP uses a deterministic balanced 16-way grouped reconstruction.
Positions whose grouped true-token reconstruction probability satisfies
$p_{\mathrm{true}}\geq 0.99$ are treated as near-deterministic and excluded from the document score.
Over the remaining positions, Target-Only JUMP averages the entropy-centered relative token preference $q_i+H_i$.
No sigma normalization is applied in the final Target-Only JUMP score.
The Masked Score baseline uses the same grouped measurements and the same $p_{\mathrm{true}}<0.99$ support, but averages $q_i$ alone.
The reference-assisted probing-budget and clipping ablations appear in
\cref{app:k_sweep_details,app:clip_operation_ablation}.

We report full-sample point estimates on the complete member/non-member sets.
Each method is evaluated on the same 1,000 members and 1,000 non-members per domain.
Metrics are computed independently within each domain and summarized by their equal-weight six-domain mean; samples are never pooled across domains or model families for the primary results.

For confidence intervals, we use class-stratified bootstrap resampling.
Within each domain, the 1,000 members and 1,000 non-members are independently resampled with replacement.
For the target-only macro confidence intervals, this procedure is repeated for 2,000 bootstrap replicates, and the six domain-level metrics are averaged replicate-wise before computing percentile 95\% intervals.
Reference-assisted confidence intervals are described in \cref{app:bootstrap_ci}.

\subsection{Target-Only JUMP Protocol}
\label[appendix]{app:reference_free_protocol}

After removing padding and structural special tokens, we assign the remaining valid positions to a deterministic balanced 16-way partition
\begin{equation*}
    \mathcal{P}(x)=\{S_1,\ldots,S_{16}\},
\end{equation*}
where group sizes differ by at most one.
Each valid position belongs to exactly one group, and each group is masked once.
Thus, every valid position is evaluated in exactly one grouped reconstruction context.

For a position $i\in S_{g(i)}$, the corresponding grouped reconstruction provides the true-token log-probability
\begin{equation*}
    q_i(S_{g(i)})
    =
    \log p_{M_{\mathrm{tgt}}}
    \left(
        x_i \mid x_{\setminus S_{g(i)}}
    \right),
\end{equation*}
and the predictive entropy
\begin{equation*}
    H_i(S_{g(i)})
    =
    -
    \sum_{v\in\mathcal{A}}
    p_{M_{\mathrm{tgt}}}
    \left(
        v \mid x_{\setminus S_{g(i)}}
    \right)
    \log
    p_{M_{\mathrm{tgt}}}
    \left(
        v \mid x_{\setminus S_{g(i)}}
    \right).
\end{equation*}
We write the corresponding grouped true-token reconstruction probability as
\begin{equation*}
    p_i(S_{g(i)})
    =
    \exp\left(q_i(S_{g(i)})\right).
\end{equation*}

The 16-way partition yields 16 sequence-equivalent masked reconstruction contexts per document.
In our implementation, eight groups are processed together in each batch, resulting in two physical batched model invocations.
We report the 16 evaluated contexts as the attack NFE rather than treating physical batching as a reduction in the number of reconstruction contexts.

We exclude near-deterministic positions and define the retained set as
\begin{equation*}
    \mathcal{R}(x)
    =
    \left\{
        i\in V(x)
        :
        p_i(S_{g(i)}) < 0.99
    \right\},
\end{equation*}
where $V(x)$ denotes the set of valid positions.
The same threshold is used for every domain and both model families.

Target-Only JUMP computes
\begin{equation*}
    g_{\mathrm{TO\text{-}JUMP}}(x;M_{\mathrm{tgt}})
    =
    \frac{1}{|\mathcal{R}(x)|}
    \sum_{i\in\mathcal{R}(x)}
    \left[
        q_i(S_{g(i)})
        +
        H_i(S_{g(i)})
    \right].
\end{equation*}
Higher values indicate stronger membership evidence.
Both filtering and scoring are obtained from the same 16 grouped reconstructions, so the entropy-centered readout introduces no additional masked contexts.
Sigma normalization is not used in the final method.

Masked Score uses exactly the same grouped measurements and retained positions but removes the entropy term:
\begin{equation*}
    g_{\mathrm{Masked}}(x;M_{\mathrm{tgt}})
    =
    \frac{1}{|\mathcal{R}(x)|}
    \sum_{i\in\mathcal{R}(x)}
    q_i(S_{g(i)}).
\end{equation*}
This comparison isolates the contribution of the entropy-centered $q+H$ readout while keeping the reconstruction contexts and filtering rule fixed.

The first three controls in \cref{tab:reference_free_results} retain their original $K=64$ definitions and are not redefined by the $p_{\mathrm{true}}<0.99$ filter.
Random samples 64 valid positions without replacement using a document-keyed hash, masks them jointly, and averages the resulting target true-token log-probabilities.
Clean Score ranks positions by the target model's clean-input true-token score, selects the 64 lowest-scoring positions, and evaluates them with a joint mask.
Ref.\ Guided uses the 64 positions selected by PRISM but removes target--reference score calibration and averages only the target true-token log-probabilities.
Thus, Masked Score and Target-Only JUMP are the two methods that share the final 16-way grouped measurements and $p_{\mathrm{true}}<0.99$ support.
\FloatBarrier

\Needspace{7\baselineskip}
\section{Bootstrap Confidence Intervals for Main Results}
\label[appendix]{app:bootstrap_ci}
We report 95\% percentile bootstrap confidence intervals for both reference-assisted and target-only JUMP on LLaDA-8B-Base and Dream-v0-Base-7B. The intervals quantify uncertainty due to the finite member/non-member evaluation sets under fixed trained models and fixed attack configurations.

For every model family, access setting, and domain, we independently resample the 1{,}000 member and 1{,}000 non-member examples with replacement for 2{,}000 class-stratified bootstrap replicates. We recompute ROC-AUC and TPR at FPR $\in\{10\%,1\%,0.1\%\}$ for each replicate and report percentile 95\% confidence intervals. We do not assume Normality of the bootstrap distributions. Domain-level metrics are never pooled across domains. For the equal-weight six-domain mean, each domain is resampled independently and the six domain-level metrics are averaged replicate-wise; the reported macro interval is the percentile interval of these replicate-wise means.

\subsection{Reference-Assisted JUMP}

\Cref{tab:bootstrap_llada_reference_assisted,tab:bootstrap_dream_reference_assisted} report per-domain uncertainty for the reference-assisted JUMP results used in the main evaluation. The point estimates reported in this appendix are the full-sample metrics computed from the same saved per-example scores used for the main results; \cref{tab:main_results} reports the corresponding values rounded to two decimal places. On LLaDA, the six-domain mean ROC-AUC is 0.9024 with a 95\% confidence interval of $[0.8972,0.9080]$. On Dream, the corresponding mean is 0.9415 with a 95\% confidence interval of $[0.9376,0.9456]$.

\begin{table}[t]
\centering
\small
\setlength{\tabcolsep}{5pt}
\caption{Per-domain JUMP results on fine-tuned LLaDA-8B-Base, with 95\% bootstrap CIs in brackets.}
\label{tab:bootstrap_llada_reference_assisted}
\label{tab:main_results_ci}
\begin{tabular}{lcccc}
\toprule
Domain & AUC & T@10 & T@1 & T@0.1 \\
\midrule
ArXiv          & 0.94 [.93,.95] & 0.84 [.80,.86] & 0.55 [.45,.60] & 0.30 [.10,.46] \\
GitHub         & 0.86 [.84,.87] & 0.66 [.62,.70] & 0.32 [.24,.44] & 0.18 [.00,.25] \\
HackerNews     & 0.76 [.74,.78] & 0.39 [.35,.45] & 0.10 [.07,.16] & 0.02 [.00,.08] \\
PubMed Central & 0.92 [.90,.93] & 0.75 [.71,.80] & 0.43 [.31,.52] & 0.12 [.04,.33] \\
Wikipedia (en) & 0.98 [.97,.99] & 0.95 [.94,.97] & 0.80 [.76,.85] & 0.67 [.61,.78] \\
Pile-CC        & 0.96 [.95,.97] & 0.90 [.88,.93] & 0.60 [.52,.67] & 0.10 [.01,.53] \\
\bottomrule
\end{tabular}
\end{table}

\begin{table}[t]
\centering
\small
\setlength{\tabcolsep}{5pt}
\caption{Per-domain JUMP results on fine-tuned Dream-v0-Base-7B, with 95\% bootstrap CIs in brackets.}
\label{tab:bootstrap_dream_reference_assisted}
\begin{tabular}{lcccc}
\toprule
Domain & AUC & T@10 & T@1 & T@0.1 \\
\midrule
ArXiv          & 0.93 [.92,.94] & 0.81 [.78,.84] & 0.52 [.39,.58] & 0.32 [.22,.40] \\
GitHub         & 0.91 [.89,.92] & 0.75 [.69,.80] & 0.36 [.28,.48] & 0.01 [.00,.29] \\
HackerNews     & 0.93 [.92,.94] & 0.81 [.77,.83] & 0.49 [.28,.58] & 0.23 [.15,.30] \\
PubMed Central & 0.95 [.94,.96] & 0.86 [.83,.89] & 0.53 [.49,.66] & 0.35 [.28,.51] \\
Wikipedia (en) & 0.96 [.95,.97] & 0.89 [.87,.92] & 0.63 [.51,.73] & 0.29 [.22,.52] \\
Pile-CC        & 0.97 [.97,.98] & 0.94 [.92,.95] & 0.77 [.66,.81] & 0.38 [.13,.68] \\
\bottomrule
\end{tabular}
\end{table}

\subsection{Target-Only JUMP}

For Target-Only JUMP, we bootstrap the final method defined in \cref{app:reference_free_protocol}. Positions with grouped true-token reconstruction probability $p_{\mathrm{true}}\geq 0.99$ are excluded, and the document score averages $q_i+H_i$ over the retained positions. No document-relative $\rho$ threshold or sigma normalization is used in these results.

\Cref{tab:bootstrap_llada_target_only,tab:bootstrap_dream_target_only} report the resulting per-domain intervals; the equal-weight six-domain intervals are summarized below. The mean ROC-AUC is 0.6095 on LLaDA with a 95\% confidence interval of $[0.5994,0.6192]$, and 0.6375 on Dream with a 95\% confidence interval of $[0.6271,0.6469]$.

Confidence intervals become wider at stricter operating points, especially TPR@0.1\% FPR. With 1{,}000 non-members per domain, an FPR of 0.1\% corresponds to approximately one false positive, so this metric is necessarily sensitive to the most extreme non-member scores. ROC-AUC and the less strict TPR operating points depend on substantially more of the score distribution and therefore have tighter intervals. These bootstrap intervals capture evaluation-set uncertainty only; they do not include variability from retraining the target model or the learned selector.
\begin{table}[H]
\centering
\small
\setlength{\tabcolsep}{5pt}
\caption{Per-domain Target-Only JUMP results on LLaDA-8B-Base, with 95\% bootstrap CIs in brackets.}
\label{tab:bootstrap_llada_target_only}
\begin{tabular}{lcccc}
\toprule
Domain & AUC & T@10 & T@1 & T@0.1 \\
\midrule
ArXiv          & 0.62 [.59,.64] & 0.21 [.17,.26] & 0.03 [.01,.06] & 0.01 [.00,.02] \\
GitHub         & 0.60 [.57,.62] & 0.21 [.17,.24] & 0.03 [.02,.05] & 0.01 [.00,.02] \\
HackerNews     & 0.57 [.54,.59] & 0.15 [.12,.19] & 0.02 [.01,.03] & 0.00 [.00,.02] \\
PubMed Central & 0.62 [.60,.64] & 0.21 [.17,.26] & 0.03 [.02,.06] & 0.01 [.00,.02] \\
Wikipedia (en) & 0.64 [.62,.66] & 0.22 [.18,.26] & 0.03 [.02,.05] & 0.01 [.00,.02] \\
Pile-CC        & 0.62 [.59,.64] & 0.20 [.16,.24] & 0.02 [.01,.04] & 0.01 [.00,.01] \\
\bottomrule
\end{tabular}
\end{table}

\begin{table}[H]
\centering
\small
\setlength{\tabcolsep}{5pt}
\caption{Per-domain Target-Only JUMP results on Dream-v0-Base-7B, with 95\% bootstrap CIs in brackets.}
\label{tab:bootstrap_dream_target_only}
\begin{tabular}{lcccc}
\toprule
Domain & AUC & T@10 & T@1 & T@0.1 \\
\midrule
ArXiv          & 0.66 [.63,.68] & 0.23 [.18,.28] & 0.04 [.02,.08] & 0.02 [.01,.03] \\
GitHub         & 0.58 [.56,.61] & 0.13 [.10,.17] & 0.01 [.00,.02] & 0.00 [.00,.01] \\
HackerNews     & 0.66 [.63,.68] & 0.22 [.19,.25] & 0.04 [.01,.07] & 0.00 [.00,.02] \\
PubMed Central & 0.64 [.61,.66] & 0.20 [.16,.25] & 0.02 [.00,.04] & 0.00 [.00,.01] \\
Wikipedia (en) & 0.65 [.63,.68] & 0.19 [.15,.25] & 0.02 [.00,.04] & 0.00 [.00,.01] \\
Pile-CC        & 0.64 [.62,.66] & 0.23 [.18,.27] & 0.04 [.02,.06] & 0.01 [.00,.02] \\
\bottomrule
\end{tabular}
\end{table}

\FloatBarrier

\Needspace{7\baselineskip}
\section{Training Exposure and Mixed-Domain Controls}
\label[appendix]{app:exposure_data_controls}
\subsection{Original-MIMIR Exposure and Memorization Controls}
\label[appendix]{app:epoch_memorization}

We train fixed one- and two-epoch LLaDA targets using the same members, non-members, optimizer, learning rate, batch size, sequence length, seed, $K$, and $\tau$ as the main evaluation.
Evaluation-based checkpoint selection is disabled for these controls.
The ``selected $\leq4$'' row uses the loss-selected checkpoints from the maximum-four-epoch training schedule.

\begin{table}[H]
\centering
\small
\setlength{\tabcolsep}{4.6pt}
\renewcommand{\arraystretch}{1.08}
\caption{Attack performance as fine-tuning exposure increases. Values are six-domain means.}
\label{tab:epoch_attack}
\begin{tabular}{lccccc}
\toprule
Exposure & Loss AUC & ZLIB AUC & SAMA AUC & \method{} AUC & \method{} TPR@1\% \\
\midrule
Epoch 1 & 0.5020 & 0.5055 & 0.5137 & \textbf{0.5263} & 0.0127 \\
Epoch 2 & 0.5056 & 0.5091 & 0.6600 & \textbf{0.7316} & 0.0942 \\
Selected $\leq4$ & 0.5273 & 0.5311 & 0.8193 & \textbf{0.9024} & 0.4663 \\
\bottomrule
\end{tabular}
\end{table}

\begin{table}[H]
\centering
\small
\setlength{\tabcolsep}{6.0pt}
\renewcommand{\arraystretch}{1.08}
\caption{Member/non-member reconstruction losses for the same exposure controls. ``Held-out gain'' is the reduction in non-member loss relative to the base model.}
\label{tab:epoch_utility}
\begin{tabular}{lcccc}
\toprule
Exposure & Member loss & Non-member loss & Gap & Held-out gain \\
\midrule
Epoch 1 & 2.4271 & 2.4432 & 0.0161 & 0.0020 \\
Epoch 2 & 2.3878 & 2.4179 & 0.0301 & 0.0273 \\
Selected $\leq4$ & 2.2604 & 2.3453 & 0.0850 & 0.0998 \\
\bottomrule
\end{tabular}
\end{table}

As shown in \cref{tab:epoch_attack,tab:epoch_utility}, absolute leakage and the member/non-member reconstruction-loss gap grow with exposure.
At one epoch, all attacks are close to chance; at two epochs, JUMP already separates members more strongly than SAMA ($0.7316$ vs. $0.6600$ AUC); the gap is largest for the selected $\leq4$ targets.
At the same time, held-out reconstruction improves rather than collapses.
Thus, the main signal is not simply a consequence of a target that fits members while losing all domain utility; it reflects a token-local advantage that becomes more consistent with exposure.

\subsection{Mixed-Domain Fine-Tuning}
\label[appendix]{app:mixed_domain}

We concatenate 1,000 members from each original MIMIR domain and fine-tune one LLaDA-8B target for one epoch on 6,000 members.
Evaluation remains within domain: each domain's members are compared with its own 1,000 disjoint non-members, and we report the equal-weight mean of the six domain metrics.
This experiment jointly changes corpus size, domain mixture, and optimizer-step count; it is a heterogeneous-target control, not a pure dataset-size ablation.

\begin{table}[ht]
\centering
\small
\setlength{\tabcolsep}{4pt}
\caption{Mixed six-domain target. The primary result is the equal-weight mean over six within-domain ROCs.}
\label{tab:mixed_domain}
\begin{tabular}{lrrrrr}
\toprule
Method & NFE & AUC & TPR@10\% & TPR@1\% & TPR@0.1\% \\
\midrule
SAMA ($T=2$) & 4 & 0.6283 & 0.2482 & 0.0515 & 0.0197 \\
SAMA ($T=16$) & 32 & 0.7116 & 0.3660 & 0.1265 & 0.0510 \\
\rowcolor{oursrow}
\method{} & 3 & \textbf{0.7466} & \textbf{0.4192} & \textbf{0.1450} & \textbf{0.0572} \\
\bottomrule
\end{tabular}
\end{table}

JUMP is strongest at all reported operating points despite using fewer forwards.
For completeness, a single pooled ROC over all 6,000 members and 6,000 non-members gives AUC $0.7199$.
We treat this as a secondary diagnostic because a pooled threshold also measures cross-domain score calibration and should not replace the primary six-domain mean.

\FloatBarrier

\Needspace{7\baselineskip}
\section{Reference-Assisted JUMP Ablations}
\label[appendix]{app:jump_ablations}
\label[appendix]{app:selector_scoring_ablations}
\label[appendix]{app:clip_and_budget}
\label[appendix]{app:runtime_details}
\subsection{Selector-Backbone Transfer}
\label[appendix]{app:reference_selector_robustness}

The exact-base model plays two distinct roles in the default attack: it calibrates the final gap between the target and reference and supplies representations to the learned selector.
To test selector dependence without changing calibration, we fix the fine-tuned LLaDA-8B target and its exact pre-fine-tuning scoring reference, then change only the model used to select positions.

\begin{table}[ht]
\centering
\small
\setlength{\tabcolsep}{4pt}
\caption{Selector-backbone transfer with the target and exact-base scoring pair fixed.}
\label{tab:selector_transfer}
\begin{tabular}{lrrrr}
\toprule
Learned selector & AUC & TPR@10\% & TPR@1\% & TPR@0.1\% \\
\midrule
LLaDA-8B-Base & 0.9024 & 0.7500 & 0.4663 & 0.2320 \\
Distilled 5.36B & 0.8966 & 0.7245 & 0.4090 & 0.1885 \\
LLaDA-1.5 (8B-class) & 0.9035 & 0.7470 & 0.4548 & 0.3152 \\
LLaDA-8B-Instruct & 0.9040 & 0.7470 & 0.4655 & 0.2550 \\
\bottomrule
\end{tabular}
\end{table}

Mean AUC varies by less than 0.008 across the compatible selector backbones.
This indicates that uncertainty-guided position selection is not tied to one selector checkpoint when the exact target/base scoring pair is retained.
Because the distilled 5.36B checkpoint also differs in distillation, pruning, and training history, it serves as a robustness control rather than a size-scaling experiment.

\subsection{Non-Exact Scoring References}
\label[appendix]{app:reference_mismatch}

Changing the scoring reference is a different experiment.
For a non-exact reference $M_R$,
\begin{equation*}
\log p_{\Mtgt}-\log p_{M_R}
=
(\log p_{\Mtgt}-\log p_{\Mbase})
+
(\log p_{\Mbase}-\log p_{M_R}),
\end{equation*}
so checkpoint mismatch is added directly to the fine-tuning-induced signal.
We fix the target and exact 8B learned selector, then replace only the model used in the final score.

\begin{table}[ht]
\centering
\small
\setlength{\tabcolsep}{4pt}
\caption{Diagnostic results with non-exact scoring references.}
\label{tab:reference_mismatch}
\begin{tabular}{lrrrr}
\toprule
Scoring reference & AUC & TPR@10\% & TPR@1\% & TPR@0.1\% \\
\midrule
Exact pre-FT LLaDA-8B & 0.9024 & 0.7500 & 0.4663 & 0.2320 \\
Distilled 5.36B & 0.6052 & 0.2137 & 0.0233 & 0.0067 \\
LLaDA-1.5 & 0.6949 & 0.3422 & 0.0666 & 0.0182 \\
LLaDA-8B-Instruct & 0.6953 & 0.3448 & 0.0670 & 0.0190 \\
\bottomrule
\end{tabular}
\end{table}

The non-exact models preserve some above-chance ranking information but sharply weaken calibration, especially at strict FPRs.
This result supports the stated scope: JUMP is designed for an exact before/after fine-tuning comparison.
Unknown-lineage attacks require an additional cross-model calibration strategy; simply substituting an approximate checkpoint is not a like-for-like implementation of the proposed score.

\subsection{Cross-Family Selector and Joint-Probing Controls}
\label[appendix]{app:selector_controls}

We compare the same selector and joint-probing controls on LLaDA and Dream.
For all joint-probing variants, we fix $K=64$, the clipping threshold
$\tau=\log 1.5$, and the clipped target-to-reference score; only the position
selection strategy changes.  Random Joint samples positions uniformly,
Entropy Top selects positions with high reference uncertainty, RareToken uses
C4 token frequencies, and JUMP uses the learned PRISM selector.  SAMA
($T=16$) is included as the previous dLLM baseline.

\begin{table}[H]
\centering
\small
\setlength{\tabcolsep}{4.3pt}
\renewcommand{\arraystretch}{1.05}
\caption{Selector and joint-probing controls across model families (six-domain mean). Joint variants use $K=64$ and $\tau=\log 1.5$; RareToken uses C4 frequencies.}
\label{tab:cross_family_selector_controls}
\begin{tabular}{llrrrr}
\toprule
Family & Method & ROC-AUC & TPR@10\% & TPR@1\% & TPR@0.1\% \\
\midrule
\multirow{5}{*}{LLaDA}
& Random Joint & 0.8173 & 0.5297 & 0.2073 & 0.0890 \\
& Entropy Top & 0.8867 & 0.7043 & 0.3650 & 0.1453 \\
& RareToken (C4-freq) & 0.8235 & 0.5489 & 0.2078 & 0.0394 \\
& SAMA ($T=16$) & 0.8193 & 0.5393 & 0.2625 & 0.1220 \\
\rowcolor{oursrow}
& \method{} & \textbf{0.9024} & \textbf{0.7500} & \textbf{0.4663} & \textbf{0.2320} \\
\midrule
\multirow{5}{*}{Dream}
& Random Joint & 0.8700 & 0.6337 & 0.2927 & 0.1365 \\
& Entropy Top & 0.8976 & 0.7190 & 0.3207 & 0.0958 \\
& RareToken (C4-freq) & 0.8175 & 0.5112 & 0.1468 & 0.0388 \\
& SAMA ($T=16$) & 0.8513 & 0.6122 & 0.2862 & 0.1660 \\
\rowcolor{oursrow}
& \method{} & \textbf{0.9415} & \textbf{0.8438} & \textbf{0.5505} & \textbf{0.2653} \\
\bottomrule
\end{tabular}
\end{table}

Across both model families, a single joint mask already yields useful signal,
but the choice of positions matters substantially.  Random Joint and RareToken
are consistently weaker, while uncertainty-based selection improves the
attack.  Learned JUMP gives the strongest mean result on both families,
supporting the same position-selection principle across the two architectures.

\subsection{Selector-Corpus Transfer}
\label[appendix]{app:selector_corpus_transfer}

We train PRISM selectors on matched 200k-example subsets of
C4~\citep{raffel2020exploring},
SlimPajama~\citep{soboleva2023slimpajama}, and
FineWeb~\citep{penedo2024fineweb}, and keep the downstream
target/reference scoring pipeline fixed.

\begin{table}[H]
\centering
\small
\caption{Six-domain mean AUC for matched selector-training corpora.}
\label{tab:selector_corpus_transfer}
\begin{tabular}{lr}
\toprule
Selector corpus & Mean ROC-AUC \\
\midrule
C4 & 0.9024 \\
SlimPajama & 0.8983 \\
FineWeb & 0.8972 \\
\bottomrule
\end{tabular}
\end{table}

\begin{figure}[htbp]
    \centering
    \includegraphics[width=0.56\linewidth]{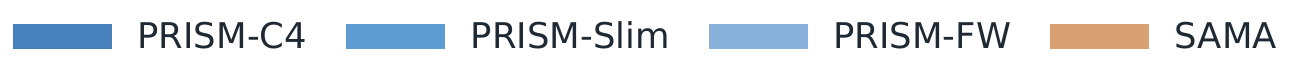}\\[-0.35em]
    \includegraphics[width=0.94\linewidth]{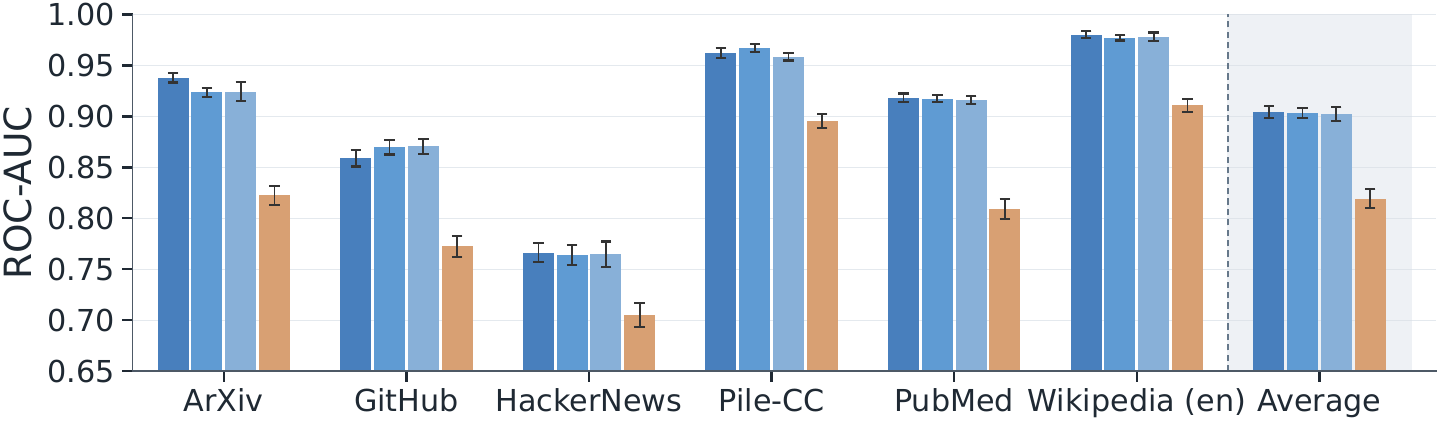}
    \caption{\textbf{Selector-corpus transfer.} Domain-wise ROC-AUC and the
    six-domain mean for PRISM selectors trained on three generic corpora.  SAMA
    is shown as the reference-assisted dLLM baseline.}
    \label{fig:selector_corpus_transfer}
\end{figure}

The C4, SlimPajama, and FineWeb selectors obtain six-domain mean AUCs of
$0.9024$, $0.8983$, and $0.8972$, respectively.  The $0.0052$ spread in
\cref{fig:selector_corpus_transfer} indicates that the learned selector is not
strongly tied to one generic corpus.  Specialized terminology, identifiers,
and numbers can still be ranked as difficult under a generic reference, so
selector transfer does not require domain-specific membership examples.  Even
the weakest of the three selectors remains $0.0779$ AUC above SAMA.

\FloatBarrier

\subsection{Low-Confidence Tokens Are a Localization Prior}
\label[appendix]{app:low_confidence_gap}

Reference low confidence does not guarantee a large improvement between the target and reference at every selected token.
A token may remain intrinsically difficult after fine-tuning, or the target may learn a domain-level pattern without memorizing that specific occurrence.
Across 256k selected token rows from the six original MIMIR domains, 23.7\% have $|\Delta_i|<0.05$.
Nevertheless, the aggregate gap between members and non-members is positive in all six domains.
This supports the intended interpretation of PRISM: it raises the density of informative probes, and the sequence-level statistic obtains power by combining many small, directionally consistent advantages.

\subsection{Rarity and Selector Overlap}
\label[appendix]{app:selector_diagnostics}

To avoid transductive use of the evaluation pool, RareToken uses an external unigram frequency table computed from 30,000 C4 sequences.
It selects the $K$ valid positions with the lowest external token frequencies and then applies the same joint masking and clipped scoring difference between the target and reference as JUMP.
The frequency table is computed entirely from external C4 text and uses no evaluation examples or membership labels.

\begin{table}[ht]
\centering
\small
\setlength{\tabcolsep}{4.5pt}
\caption{LLaDA selector diagnostics (six-domain mean); RareToken uses C4 frequencies.}
\label{tab:selector_diagnostic_summary}
\begin{tabular}{lrrrr}
\toprule
Method & ROC-AUC & TPR@10\% & TPR@1\% & TPR@0.1\% \\
\midrule
RareToken (C4-freq) & 0.8235 & 0.5489 & 0.2078 & 0.0394 \\
SAMA ($T=16$) & 0.8193 & 0.5393 & 0.2625 & 0.1220 \\
\rowcolor{oursrow}
\method{} & \textbf{0.9024} & \textbf{0.7500} & \textbf{0.4663} & \textbf{0.2320} \\
\bottomrule
\end{tabular}
\end{table}

\cref{tab:selector_diagnostic_summary} shows that external-corpus rarity is
weaker than JUMP at every operating point, especially at TPR@0.1\% FPR.  SAMA
provides the prior dLLM baseline.  The overlap analysis below compares JUMP
with random, entropy-based, and rarity-based position sets.

\begin{figure}[ht]
\centering
\includegraphics[width=0.88\linewidth]{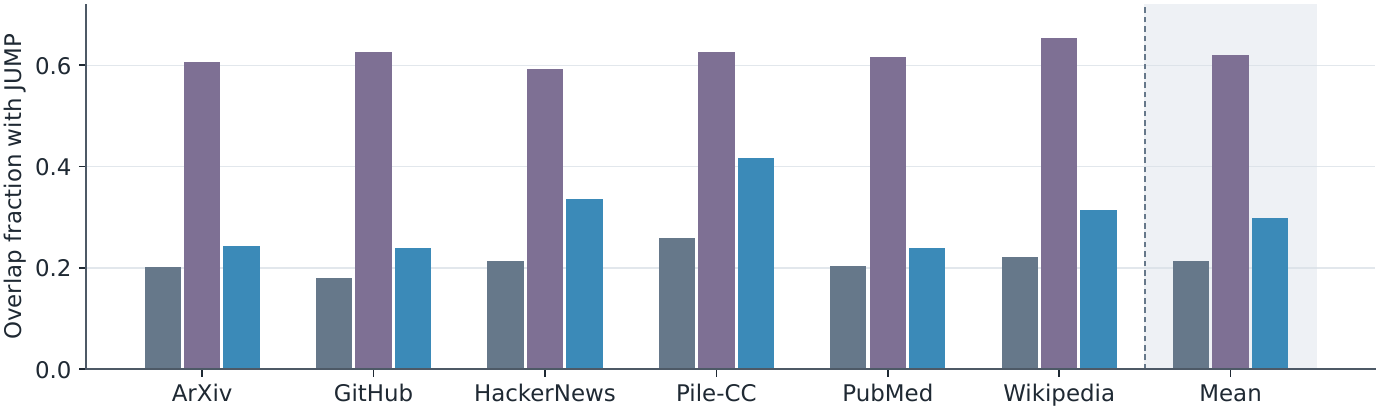}
\vspace{0.3em}
\includegraphics[width=0.88\linewidth]{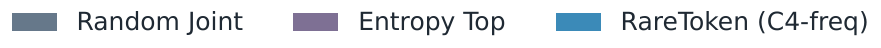}
\caption{Overlap between JUMP-selected positions and alternative selector rules. Entropy Top has the largest overlap, while RareToken and Random Joint overlap much less, indicating that the learned selector is not equivalent to a single hand-designed heuristic.}
\label{fig:selector_overlap}
\end{figure}

Averaged over domains, Entropy Top overlaps with JUMP on 62.0\% of positions,
RareToken on 29.8\%, and Random Joint on 21.3\%.
The partial overlap is consistent with PRISM learning contextual reconstruction difficulty rather than only rarity or clean-text probability.

\FloatBarrier

\subsection{Selector Training Sample-Size Ablation}
\label[appendix]{app:c4_sample_size_ablation}

We ablate the amount of C4 data used to train the PRISM head.
Using the same LLaDA backbone and selector-training recipe, we train heads on $20\mathrm{k}$, $50\mathrm{k}$, $100\mathrm{k}$, $200\mathrm{k}$, $400\mathrm{k}$, and $500\mathrm{k}$ examples.
All selectors are evaluated with the main JUMP pipeline: $K=64$, a single joint masked query, the difference between target and reference true-token log-probabilities, and symmetric clipping at $\tau=\log 1.5$.

\begin{figure}[ht]
\centering
\includegraphics[width=0.50\linewidth]{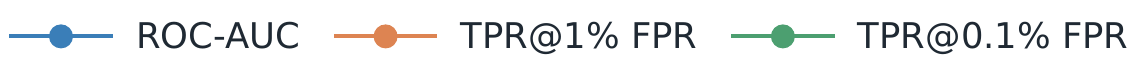}\\[-0.35em]
\begin{minipage}[t]{0.325\linewidth}
    \centering
    \includegraphics[width=\linewidth]{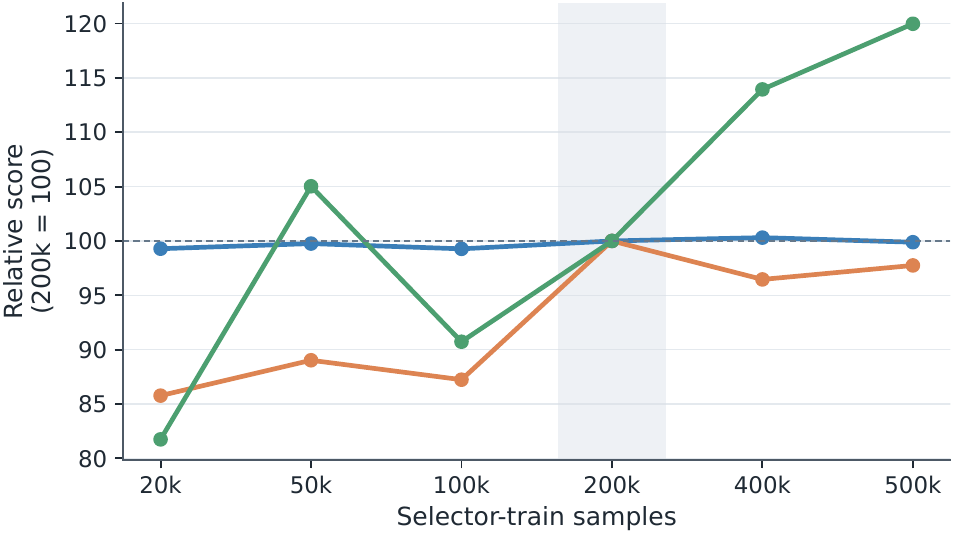}
\end{minipage}\hfill
\begin{minipage}[t]{0.325\linewidth}
    \centering
    \includegraphics[width=\linewidth]{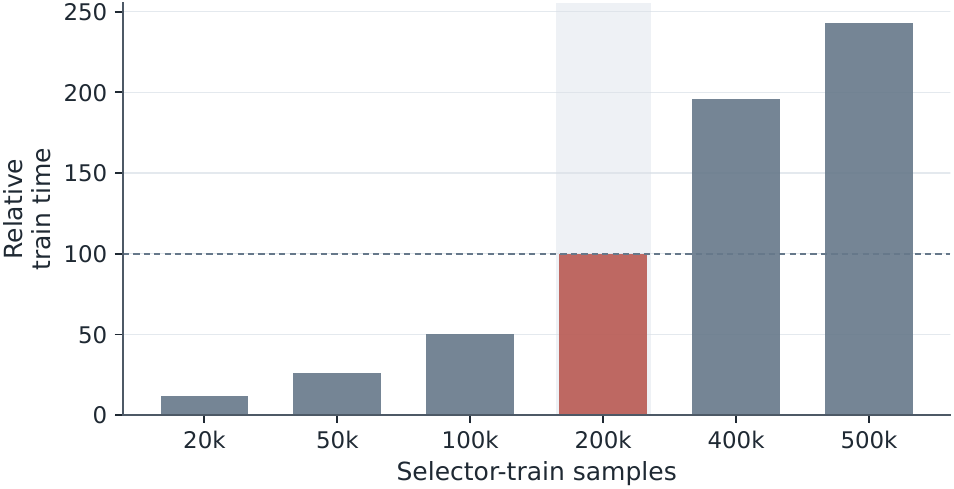}
\end{minipage}\hfill
\begin{minipage}[t]{0.325\linewidth}
    \centering
    \includegraphics[width=\linewidth]{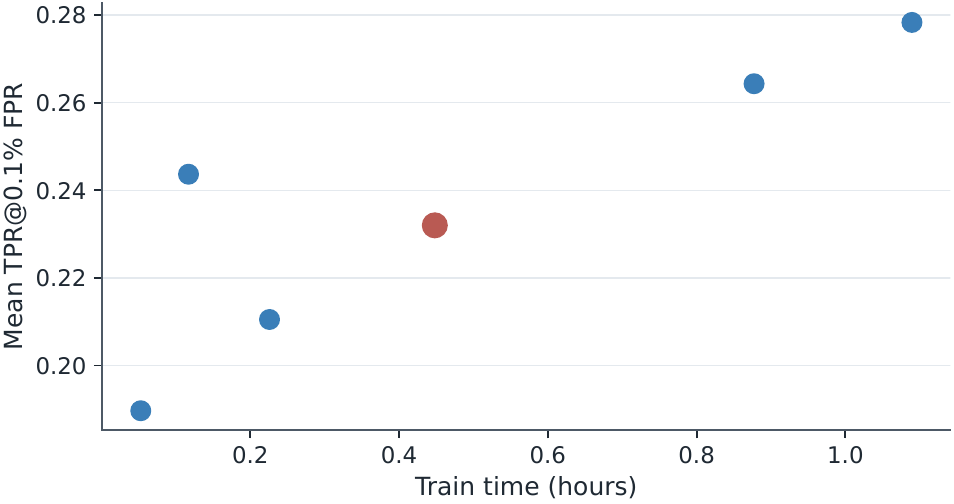}
\end{minipage}
\caption{Effect of selector-training set size.
Performance is already strong at small scales and changes only modestly beyond $200\mathrm{k}$ examples.
The $200\mathrm{k}$ selector provides a balanced operating point, while larger sets yield small metric-specific gains at higher training cost.}
\label{fig:c4_sample_size_ablation}
\end{figure}

\begin{table}[ht]
\centering
\small
\setlength{\tabcolsep}{3.8pt}
\caption{C4 selector-training sample-size ablation, averaged over six LLaDA domains.}
\label{tab:c4_sample_size_ablation}
\begin{tabular}{lrrrrr}
\toprule
C4 samples & ROC-AUC & TPR@10\% & TPR@1\% & TPR@0.1\% & Train time \\
\midrule
$20\mathrm{k}$  & 0.8960 & 0.7303 & 0.4000 & 0.1897 & 190s \\
$50\mathrm{k}$  & 0.9002 & 0.7437 & 0.4152 & 0.2437 & 421s \\
$100\mathrm{k}$ & 0.8959 & 0.7195 & 0.4068 & 0.2105 & 813s \\
\rowcolor{oursrow}
$200\mathrm{k}$ & 0.9024 & \textbf{0.7500} & \textbf{0.4663} & 0.2320 & 1{,}613s \\
$400\mathrm{k}$ & \textbf{0.9052} & 0.7477 & 0.4498 & 0.2643 & 3{,}158s \\
$500\mathrm{k}$ & 0.9014 & 0.7432 & 0.4558 & \textbf{0.2783} & 3{,}922s \\
\bottomrule
\end{tabular}
\end{table}

For reference, the main SAMA configuration uses $T=16$ ($2T=32$ target/reference forwards) and takes 1.863 seconds per sample. Its online cost reaches the default $200\mathrm{k}$ PRISM training time of 1,613 seconds after approximately 866 audited samples; over the 12,000-example suite, selector training amortizes to 0.134 seconds per sample.

\cref{tab:c4_sample_size_ablation,fig:c4_sample_size_ablation} show that selector quality is already high with $20\mathrm{k}$ examples and largely saturates around $200\mathrm{k}$.
The $200\mathrm{k}$ selector reproduces the main JUMP result and achieves the best TPR@10\% and TPR@1\% FPR among the tested budgets.
Increasing the corpus to $400\mathrm{k}$ slightly improves mean ROC-AUC, while $500\mathrm{k}$ improves TPR@0.1\% FPR; neither larger budget improves all metrics simultaneously.
We therefore use $200\mathrm{k}$ as a balanced default that avoids the approximately $1.96\times$ and $2.43\times$ training-time increases of the $400\mathrm{k}$ and $500\mathrm{k}$ selectors.

\subsection{Selected-Token Budget Sweep}
\label[appendix]{app:k_sweep_details}

We sweep $K\in\{32,64,128\}$ with $\tau=\log 1.5$ fixed, while holding the target checkpoints, selector, evaluation examples, and aggregation rule constant.

\begin{table}[H]
\centering
\small
\setlength{\tabcolsep}{5pt}
\caption{Selected-token budget sweep with $\tau=\log 1.5$, averaged over six LLaDA domains.}
\label{tab:k_sweep}
\begin{tabular}{rrrrr}
\toprule
$K$ & ROC-AUC & TPR@10\% & TPR@1\% & TPR@0.1\% \\
\midrule
32 & 0.8880 & 0.7100 & 0.3995 & 0.1798 \\
64 & \textbf{0.9024} & \textbf{0.7500} & \textbf{0.4663} & \textbf{0.2320} \\
128 & 0.8650 & 0.6525 & 0.2430 & 0.0442 \\
\bottomrule
\end{tabular}
\end{table}

\begin{figure}[H]
    \centering
    \includegraphics[width=0.46\linewidth]{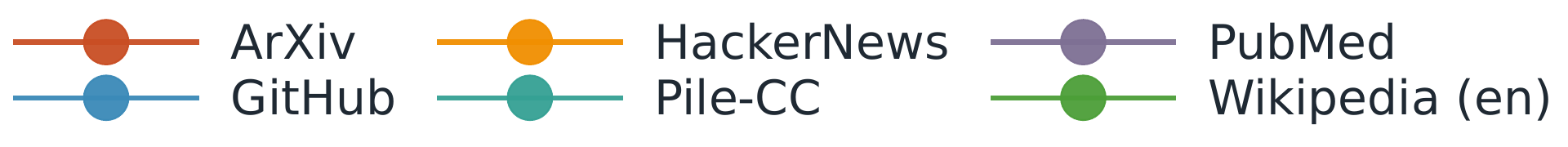}\\[-0.25em]
    \includegraphics[width=0.64\linewidth]{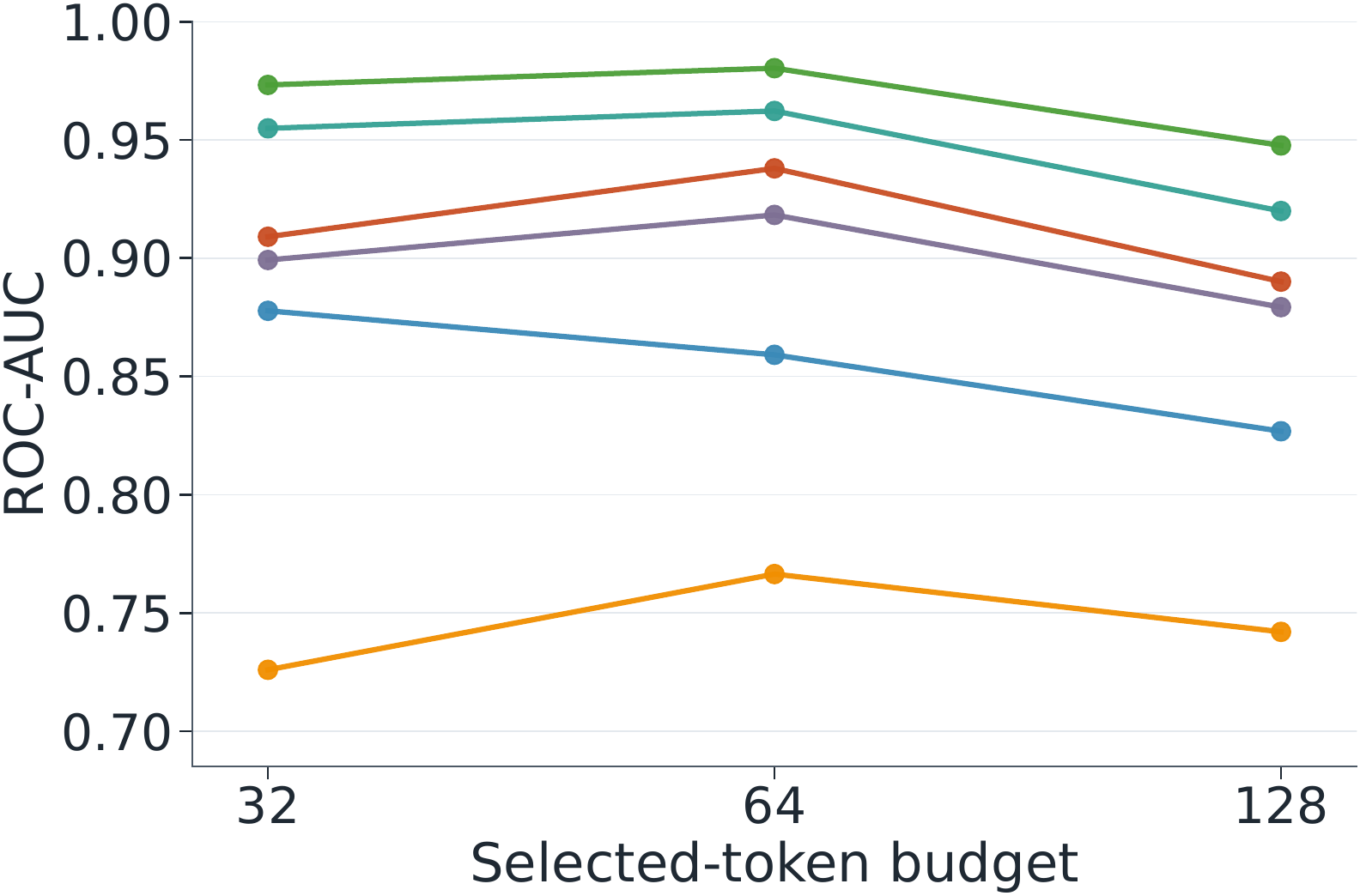}
    \caption{\textbf{Selected-token budget.} Domain-wise ROC-AUC across
    probing budgets.  The six-domain average is maximized at $K=64$.}
    \label{fig:k_sweep_auc}
\end{figure}

\begin{figure}[H]
    \centering
    \begin{minipage}[t]{0.325\linewidth}
        \centering
        \includegraphics[width=\linewidth]{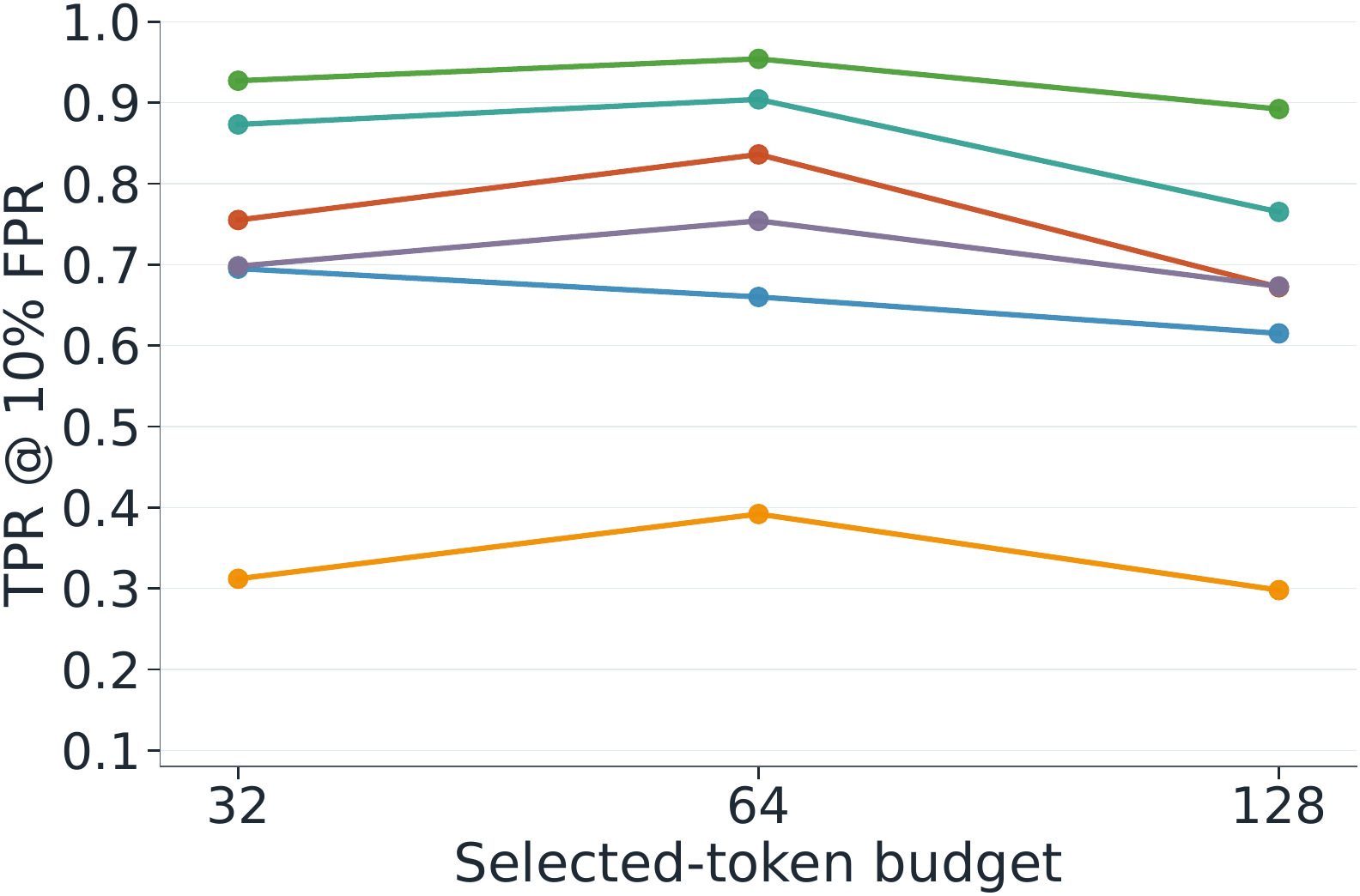}
    \end{minipage}\hfill
    \begin{minipage}[t]{0.325\linewidth}
        \centering
        \includegraphics[width=\linewidth]{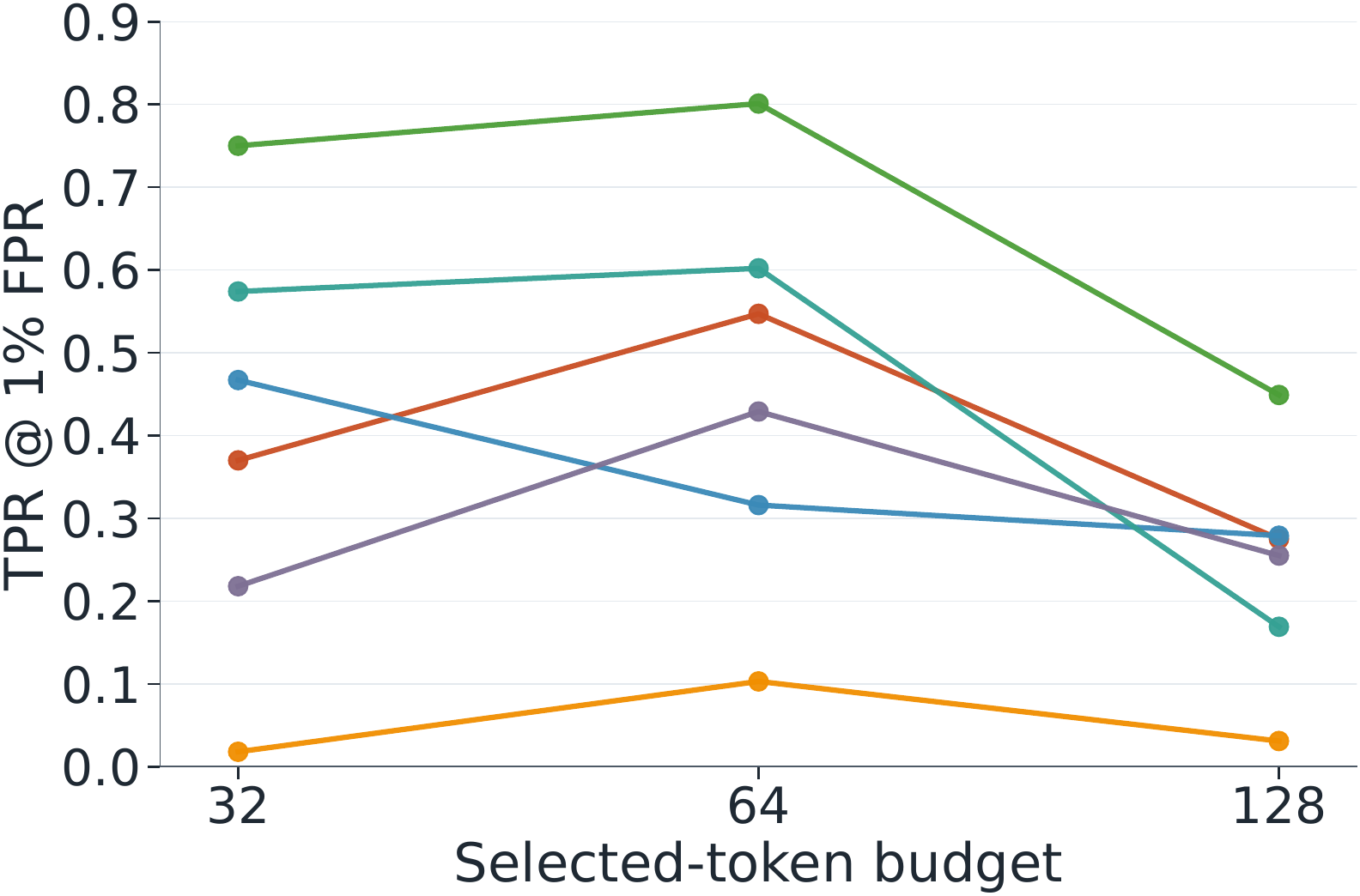}
    \end{minipage}\hfill
    \begin{minipage}[t]{0.325\linewidth}
        \centering
        \includegraphics[width=\linewidth]{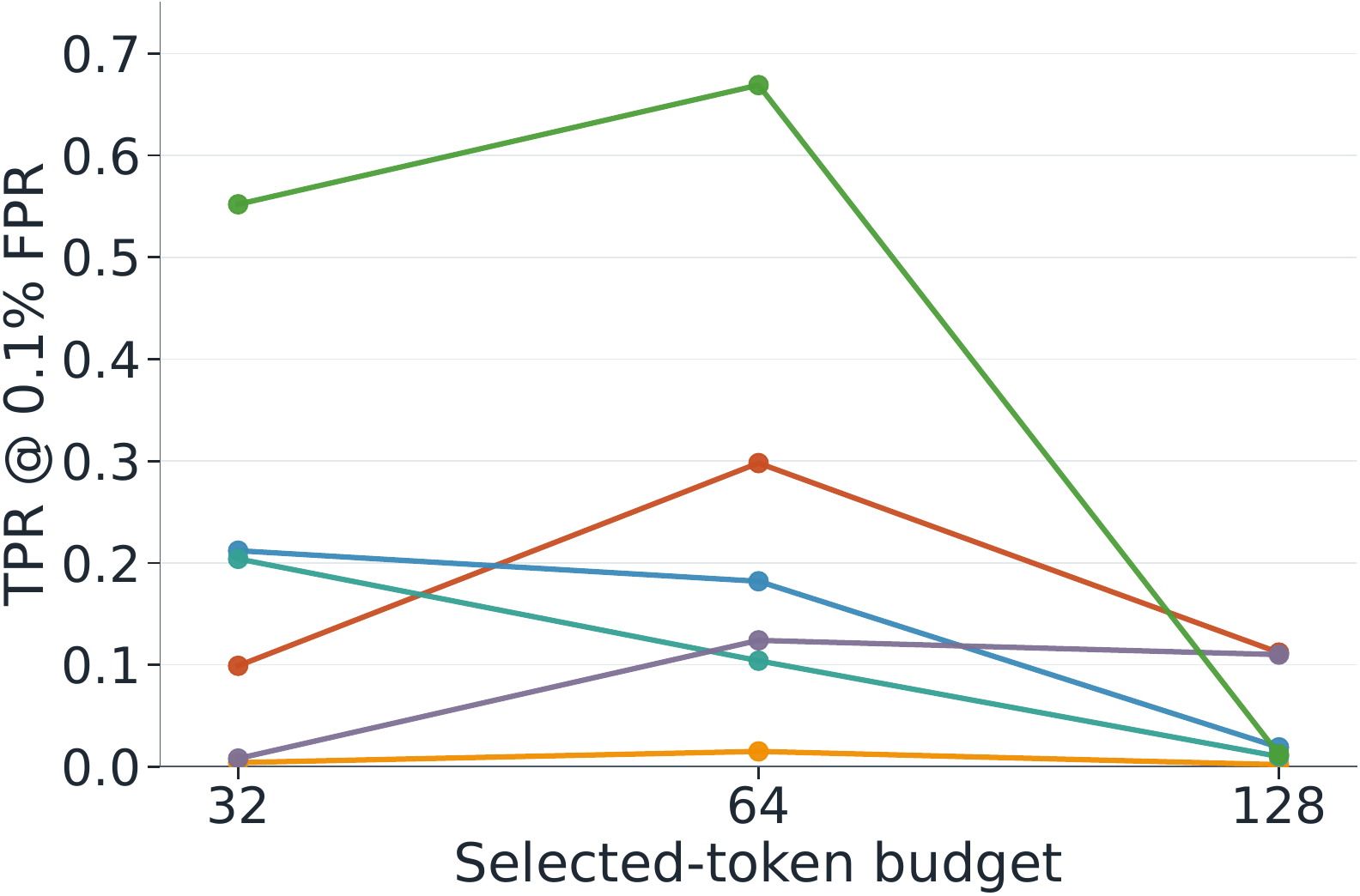}
    \end{minipage}\\[-0.15em]
    \includegraphics[width=0.62\linewidth]{figures/k_sweep_legend.pdf}
    \caption{Detailed selected-token budget sweep at low-FPR operating points. The same domain ordering is preserved across panels, and $K=64$ gives the strongest mean operating point.}
    \label{fig:k_sweep_lowfpr}
\end{figure}

With $K=32$, the statistic averages fewer local reconstruction advantages;
with $K=128$, less informative positions dilute the selected set.  Both the
domain-wise ROC-AUC curves in \cref{fig:k_sweep_auc} and the strict operating
points in \cref{fig:k_sweep_lowfpr} support $K=64$.  Increasing $K$ does not
change the number of model forwards because all retained positions are scored
in the same joint masked input.

\FloatBarrier

\subsection{Clipping Operation Ablation}
\label[appendix]{app:clip_operation_ablation}

\Cref{tab:clip_ablation_full} reports the aggregate clipping ablation.
The selector is fixed to the PRISM-C4 head, and both readouts use the same selected token positions.
The unclipped variant averages raw target/reference gaps; the clipped variant caps each token contribution before averaging.

\begin{table}[htbp]
\centering
\small
\setlength{\tabcolsep}{5pt}
\caption{Clipping ablation averaged over six MIMIR domains.}
\label{tab:clip_ablation_full}
\begin{tabular}{lrrrr}
\toprule
Method & ROC-AUC & TPR@10\% & TPR@1\% & TPR@0.1\% \\
\midrule
No clip mean & 0.8166 & 0.5478 & 0.2323 & 0.1003 \\
\rowcolor{oursrow}
Clipped mean & \textbf{0.9024} & \textbf{0.7500} & \textbf{0.4663} & \textbf{0.2320} \\
\bottomrule
\end{tabular}
\end{table}

\begin{figure}[H]
    \centering
    \begin{subfigure}[t]{0.48\linewidth}
        \centering
        \includegraphics[width=0.90\linewidth]{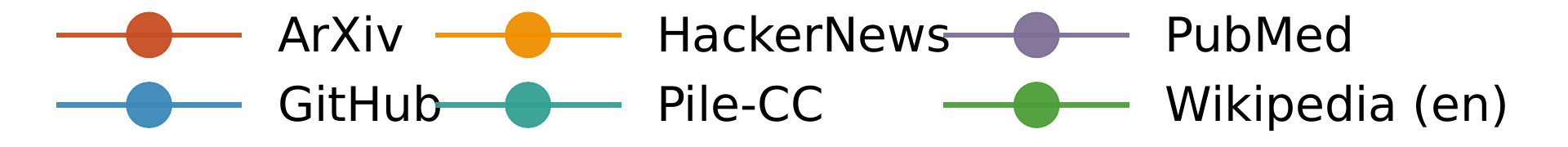}\\[-0.35em]
        \includegraphics[width=\linewidth]{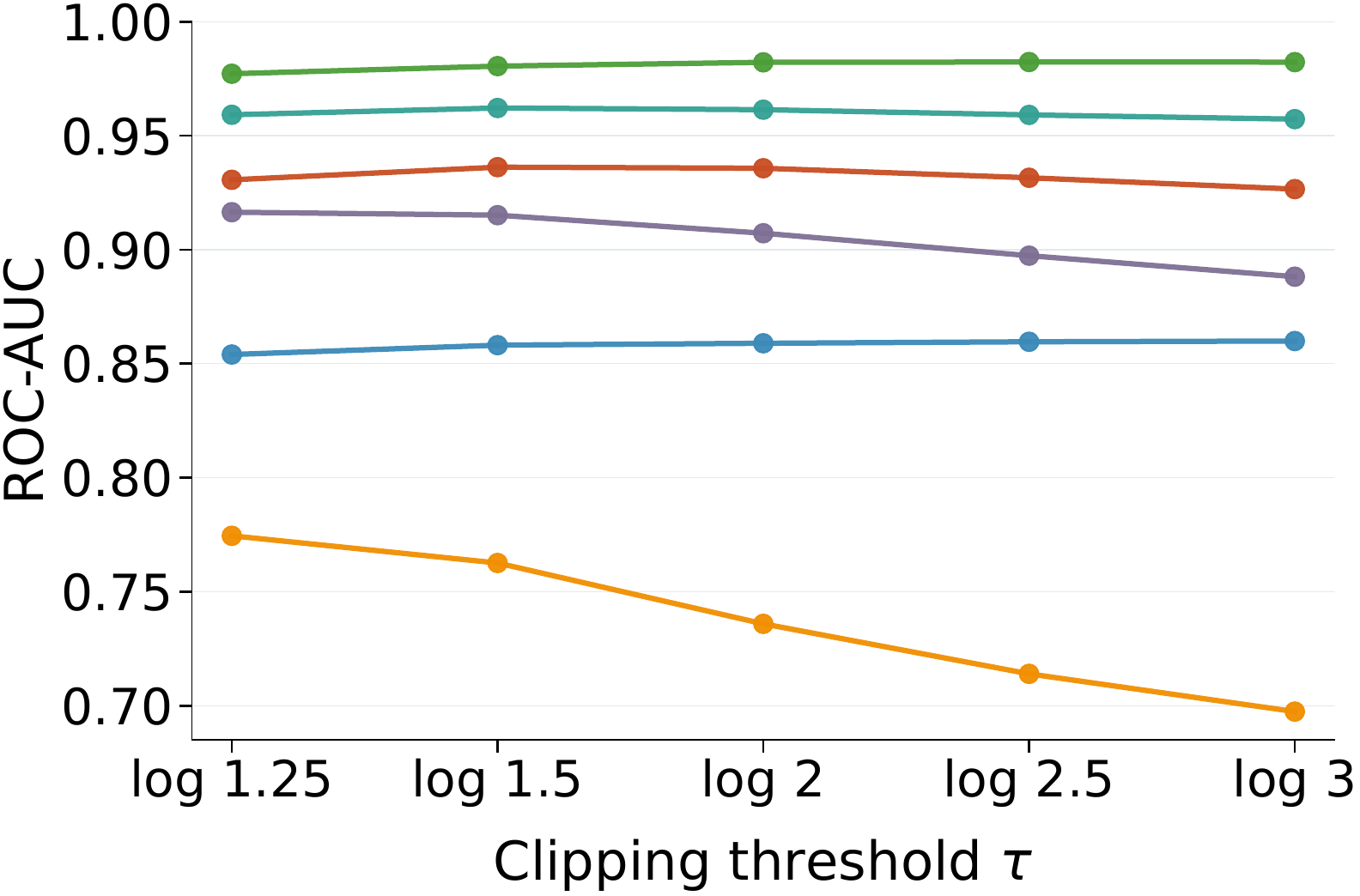}
        \caption{Threshold sensitivity.}
    \end{subfigure}
    \hfill
    \begin{subfigure}[t]{0.48\linewidth}
        \centering
        \includegraphics[width=0.90\linewidth]{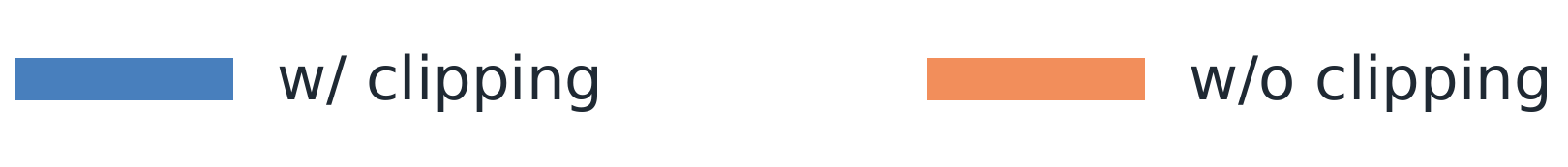}\\[-0.35em]
        \includegraphics[width=\linewidth]{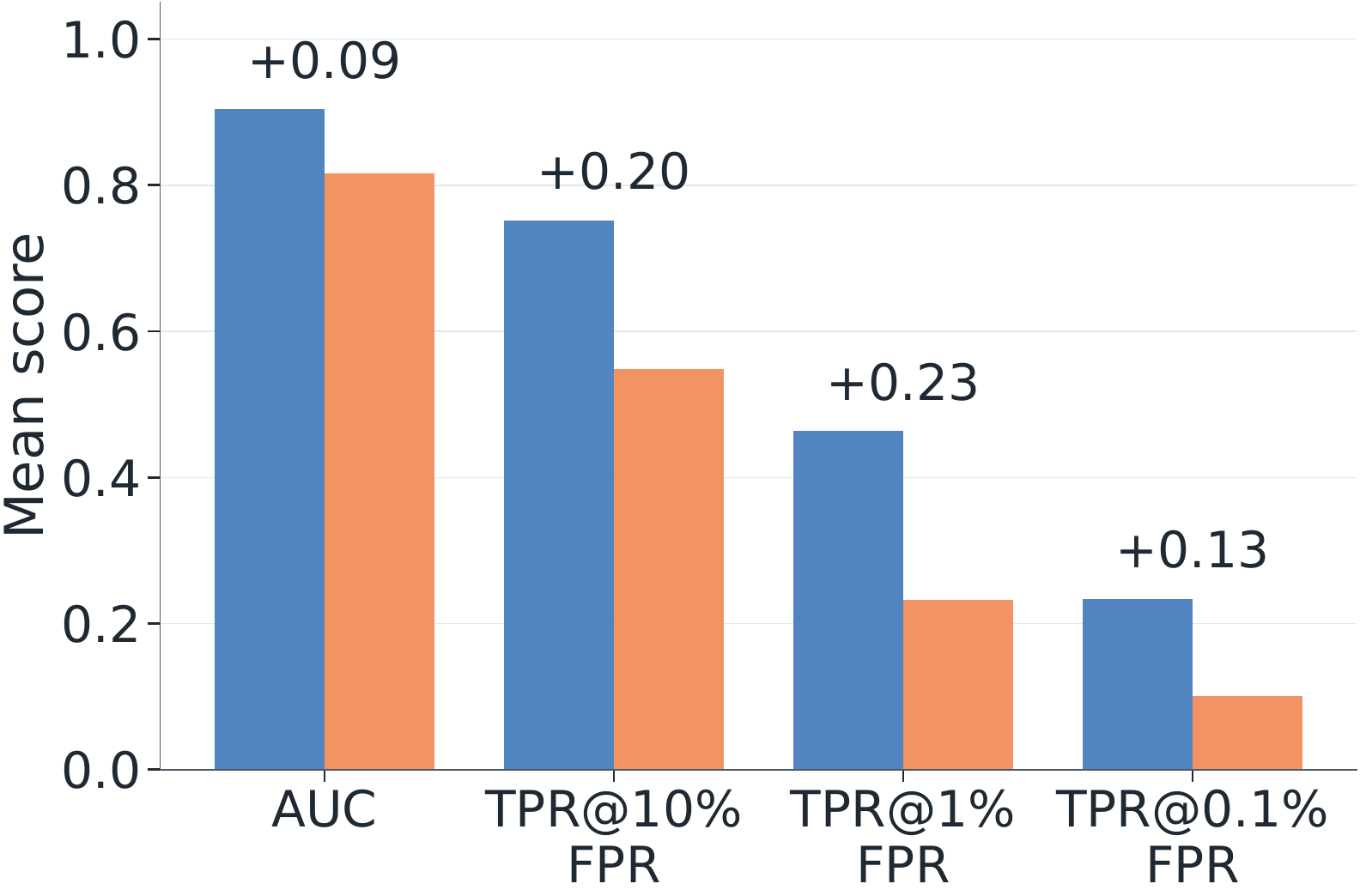}
        \caption{Clipped versus unclipped mean.}
    \end{subfigure}
    \caption{\textbf{Clipping ablation.} The left panel varies the clipping
    threshold; the right panel isolates the effect of applying clipping while
    keeping the selected positions fixed.}
    \label{fig:clipping_combined}
\end{figure}

Clipping improves every metric, with the largest relative gain at the strictest operating point.
In particular, TPR@0.1\%FPR more than doubles after clipping.
This behavior is consistent with the role of clipping as a robustness correction: selected hard positions can produce extreme target/reference gaps, and an unclipped mean can be dominated by a small number of unstable token-level contributions.
Clipping limits these isolated spikes and instead rewards sequences where the target has a stable reconstruction advantage across multiple selected positions.

\subsection{Clipping-Threshold Sweep}
\label[appendix]{app:clip_sweep}

The threshold $\tau$ controls the cap applied to each token contribution in
\cref{eq:score}.  Small values suppress token-level outliers more aggressively,
whereas large values approach the unclipped mean.  The ROC-AUC curves in
\cref{fig:clipping_combined} show that the method is not narrowly tuned to a
single threshold.  We additionally report the complete low-FPR sweep below.

\begin{figure}[H]
    \centering
    \begin{subfigure}[t]{0.32\linewidth}
        \centering
        \includegraphics[width=\linewidth]{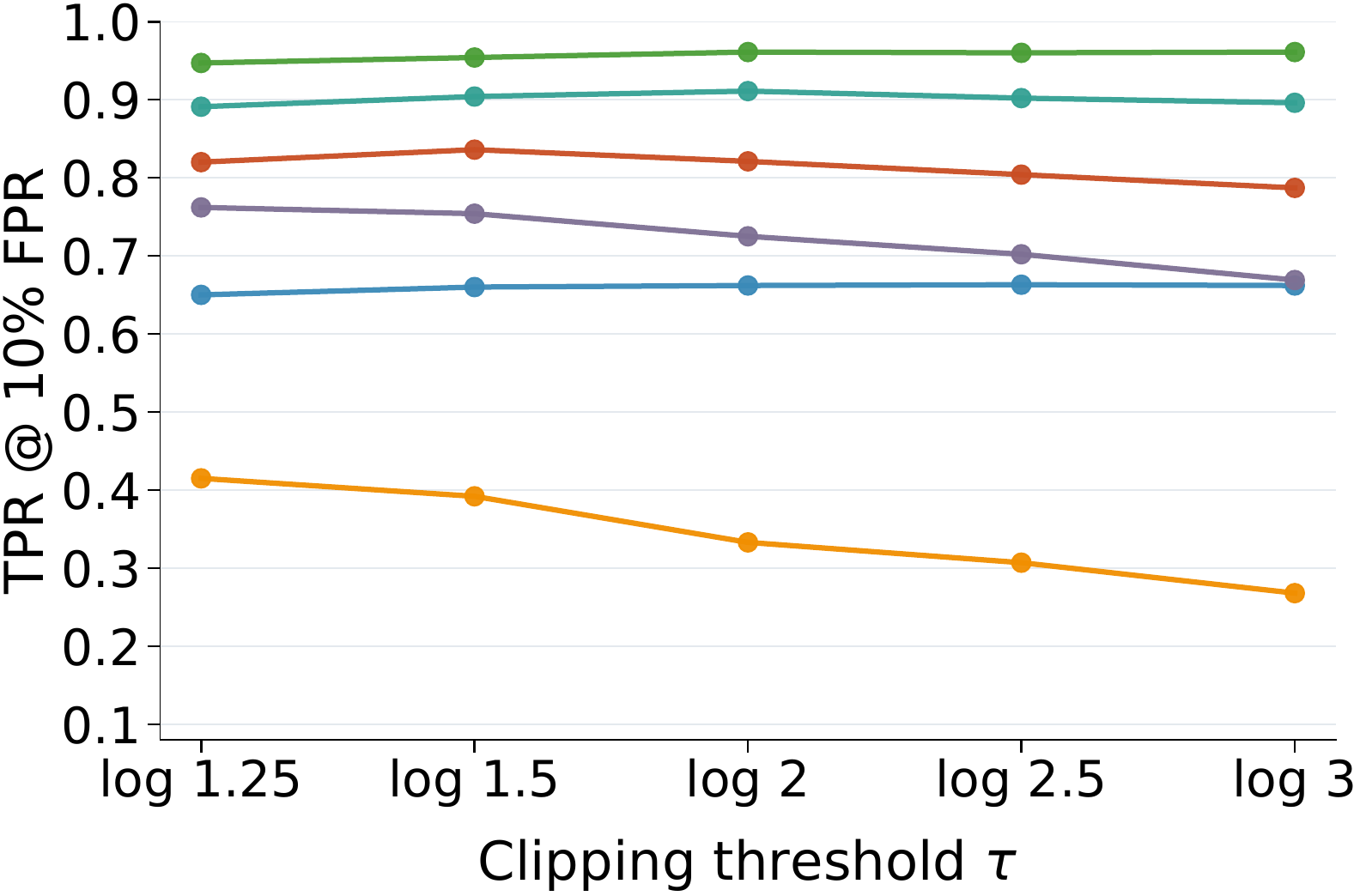}
        \caption{TPR@10\% FPR.}
        \label{fig:clip_sweep_tpr10}
    \end{subfigure}
    \hfill
    \begin{subfigure}[t]{0.32\linewidth}
        \centering
        \includegraphics[width=\linewidth]{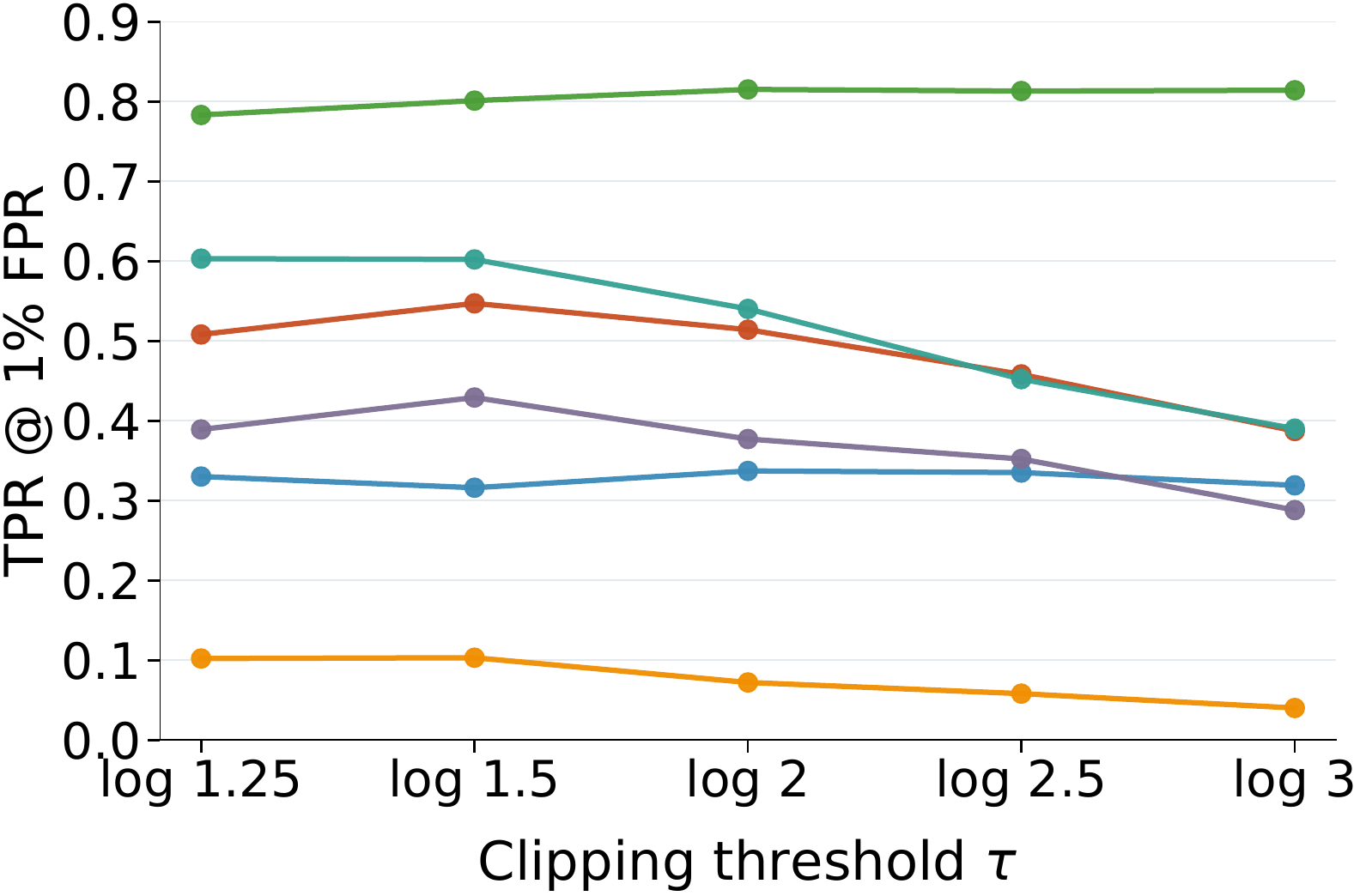}
        \caption{TPR@1\% FPR.}
        \label{fig:clip_sweep_tpr1}
    \end{subfigure}
    \hfill
    \begin{subfigure}[t]{0.32\linewidth}
        \centering
        \includegraphics[width=\linewidth]{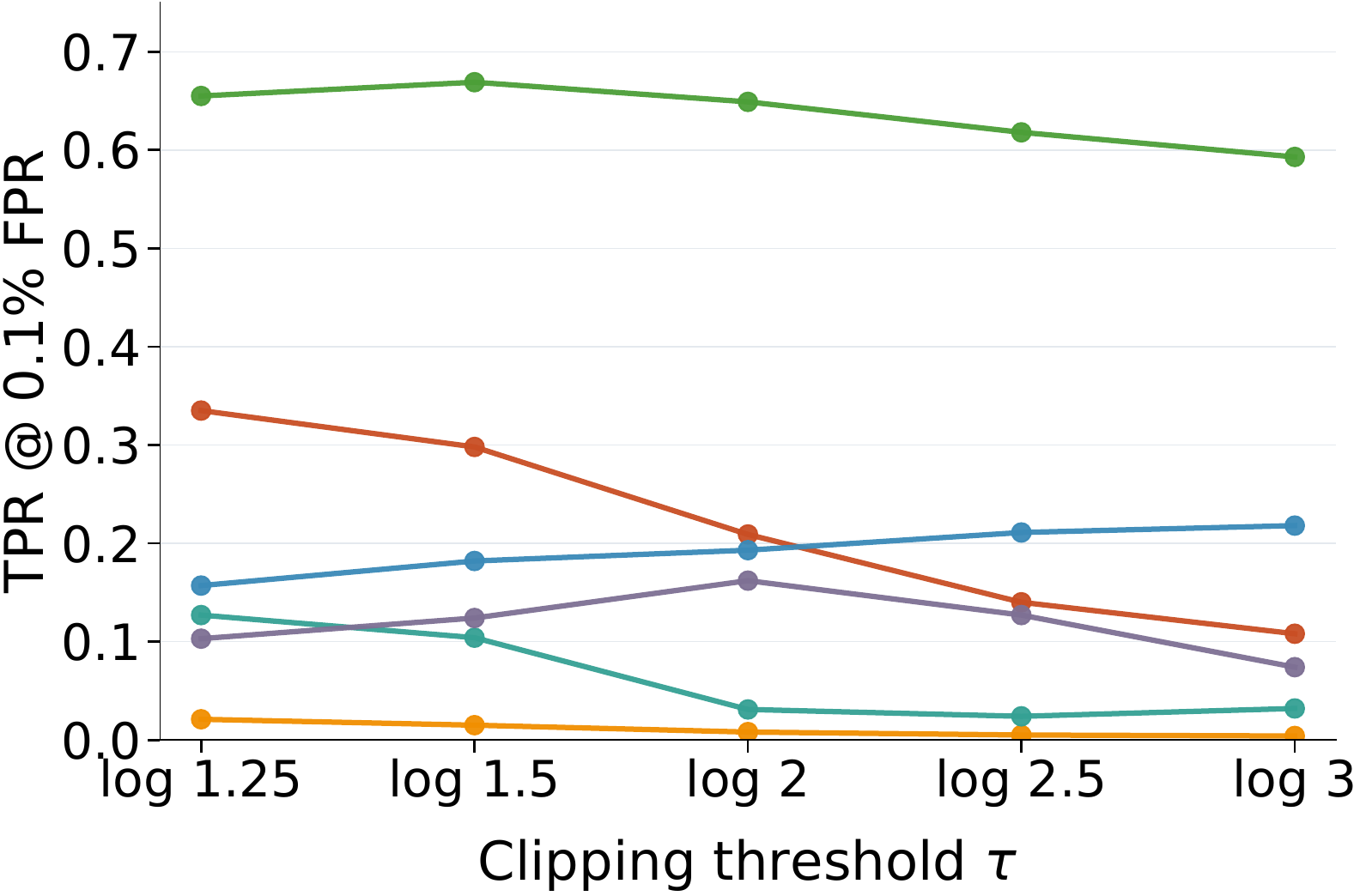}
        \caption{TPR@0.1\% FPR.}
        \label{fig:clip_sweep_tpr01}
    \end{subfigure}

    \vspace{0.35em}
    \includegraphics[width=0.62\linewidth]{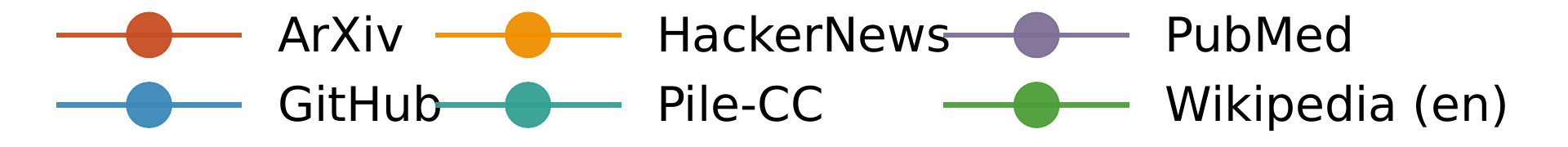}

    \caption{\textbf{Clipping-threshold sensitivity at low FPR.} The legend is
    shared across panels.  Although individual domains sometimes favor a
    neighboring threshold, $\tau=\log 1.5$ gives the strongest overall balance
    across the three operating points.}
    \label{fig:clip_sweep_appendix}
\end{figure}

Across the full domain and metric grid, $\tau=\log 1.5$ provides the most
stable aggregate operating point.  We therefore use it for every
reference-assisted experiment rather than selecting a threshold separately for
each domain or FPR.

\subsection{Selector Training Cost}
\label[appendix]{app:selector_training_runtime}

For the selector-corpus ablation in \cref{tab:selector_corpus_transfer}, all PRISM heads are trained with the same budget of 200k samples.
This matched budget isolates the effect of the selector training corpus rather than the amount of selector supervision.
The downstream attack pipeline is otherwise identical across PRISM-C4, PRISM-Slim, and PRISM-FW: each selector chooses $K=64$ positions, after which the same joint multimask query and clipped scoring difference between the target and reference are applied.

PRISM-head training is a one-time preprocessing cost.
In our implementation, PRISM-C4 takes 26.9 minutes of wall time, corresponding to approximately 1.8 GPU-hours on 4$\times$L40S GPUs.
PRISM-Slim takes 23.9 minutes, or approximately 1.6 GPU-hours, and PRISM-FW takes 26.8 minutes, or approximately 1.8 GPU-hours.
When amortized over the evaluation suite of six datasets with 2,000 samples each, these costs correspond to 0.134 seconds per sample for PRISM-C4, 0.120 seconds per sample for PRISM-Slim, and 0.134 seconds per sample for PRISM-FW.

\subsection{Per-Sample Runtime Accounting}
\label[appendix]{app:per_sample_runtime}

All reported runtimes are normalized as seconds per sample.
Although attacks are executed in batches for efficiency, per-sample runtime is the most direct measure of attack cost from the auditor's perspective.
This is particularly important for dLLM MIAs because attacks that repeatedly query many masked variants can become expensive even when each individual forward pass is parallelized.

\begin{table}[ht]
\centering
\small
\setlength{\tabcolsep}{5pt}
\caption{Runtime per sample in seconds. ``Head'' denotes amortized one-time selector training.}
\label{tab:app_timing}
\begin{tabular}{lrrr}
\toprule
Method & Inference & Head & Total \\
\midrule
SAMA & 1.863 & N/A & 1.863 \\
JUMP (C4) & 0.184 & 0.134 & 0.318 \\
JUMP (SlimPajama) & 0.185 & 0.120 & 0.305 \\
JUMP (FineWeb) & 0.183 & 0.134 & 0.317 \\
\bottomrule
\end{tabular}
\end{table}

Even after amortizing selector-head training, the learned-selector variants remain substantially faster than SAMA.
The single-pass design provides the computational advantage: after selection, all retained tokens are reconstructed jointly in one masked input.

\FloatBarrier

\Needspace{7\baselineskip}
\section{Role of Reference Calibration}
\label[appendix]{app:reference_calibration}
The reference model supports both position selection and token-level
calibration.  To measure the calibration contribution, we fix the
PRISM-selected $K=64$ positions and compare their target reconstruction scores
with the target-to-reference gap in \cref{eq:gap_def}.  The selected positions
and evaluation examples are identical in both cases.

\begin{figure}[htbp]
    \centering
    \includegraphics[width=0.52\linewidth]{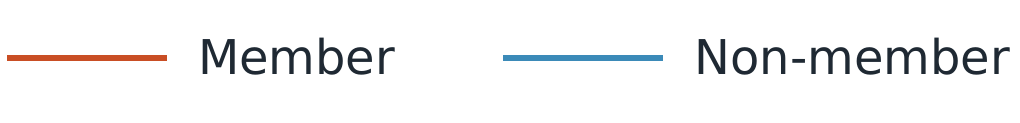}\\[0.35em]
    \includegraphics[width=0.98\linewidth]{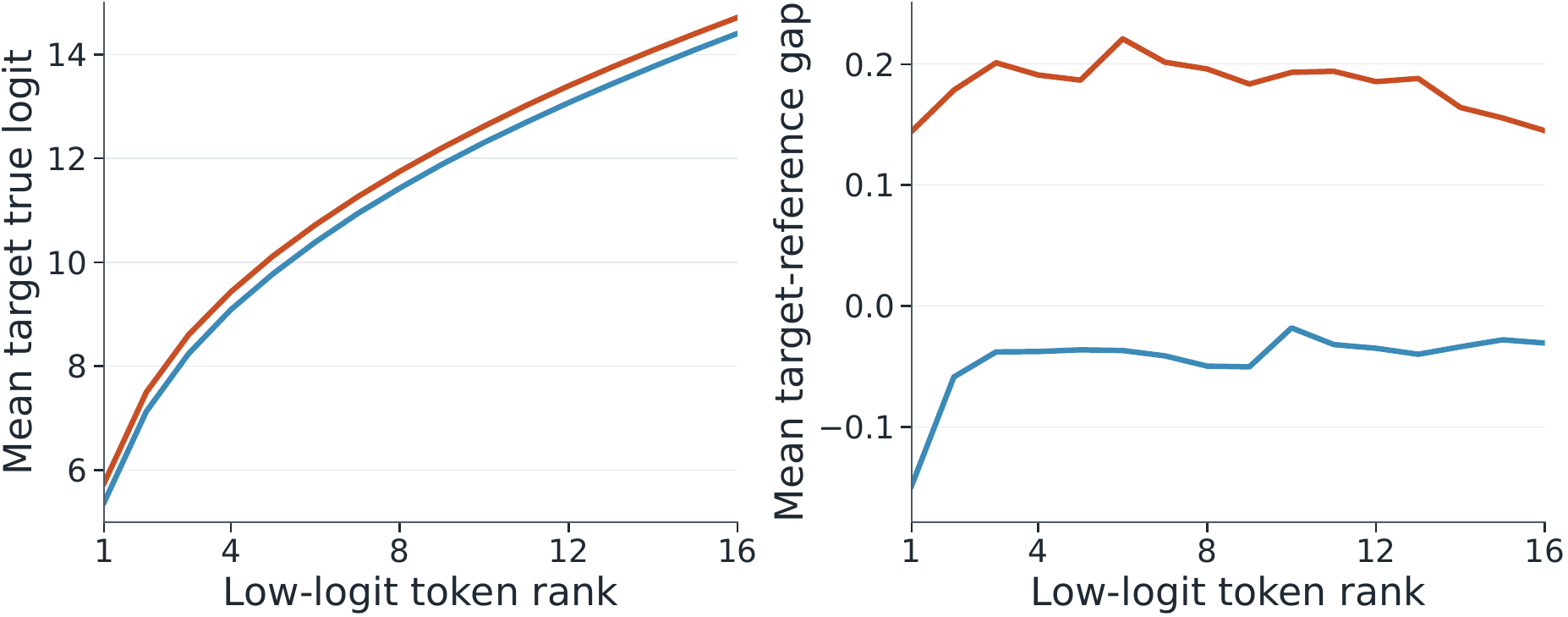}
    \caption{\textbf{Effect of reference calibration.} The left panel reports
    target true-token scores on the PRISM-selected positions.  The right panel
    evaluates the target-to-reference gap on those same positions.
    Absolute scores retain substantial token-difficulty variation, whereas the
    paired gap produces a clearer separation between members and non-members.}
    \label{fig:reference_needed}
\end{figure}

\Cref{fig:reference_needed} visualizes the token-level mechanism, and the
sequence-level ablation gives the same conclusion. With the PRISM mask set
held fixed, the target-only readout reaches mean ROC-AUC $0.5616$ and
TPR@10\% FPR $0.1462$ on LLaDA, compared with $0.9024$ and $0.7500$ for
reference-calibrated JUMP. On Dream, the corresponding values are $0.6176$
and $0.2032$, versus $0.9415$ and $0.8438$. Thus, even a learned selector does
not make an absolute target score directly comparable across documents. A low
score may reflect intrinsic ambiguity rather than non-membership, while a high
score may reflect generic predictability rather than memorization. The paired
gap addresses this ambiguity when the lineage checkpoint is available; the
entropy-centered target-only construction in \cref{sec:target_only} is used
when it is not.

The broader selector controls in \cref{tab:reference_free_results} reinforce
this separation. Random positions, Clean confidence, and Reference-guided reach LLaDA mean AUC
$0.5530$, $0.5603$, and $0.5616$, respectively; the corresponding Dream
values are $0.5899$, $0.6143$, and $0.6176$.
Low-confidence selection therefore adds signal, especially on Dream, but the
small change from target-clean to PRISM-selected target scoring confirms that
better localization alone is not a substitute for calibrating token
difficulty.

\section{Target-Only JUMP Ablations}
\label[appendix]{app:reference_free_ablations}
We evaluate the components of Target-Only JUMP using the same
1{,}000 members and 1{,}000 non-members per MIMIR domain as in the
main evaluation.
Unless otherwise stated, all analyses reuse the deterministic balanced
16-way grouped reconstructions defined in
\cref{app:reference_free_protocol}.
The threshold, readout, and normalization ablations therefore
reaggregate the same target-model outputs and introduce no additional
masked reconstruction contexts.

We focus on three questions: why near-deterministic positions should be
removed, how the entropy-centered readout behaves across domains and
model families, and whether the 16-way grouped reconstruction provides
a sufficiently accurate approximation to the ideal one-hole
measurement.

\subsection{Reconstruction Confidence and the Near-Deterministic Regime}
\label[appendix]{app:target_only_saturation_distribution}

Target-Only JUMP removes positions whose grouped true-token
reconstruction probability is at least $0.99$.
This filter is intended to exclude a saturated reconstruction regime,
rather than to select a fixed fraction of low-confidence tokens.
To characterize this regime directly, we examine the empirical
distribution of grouped true-token reconstruction probabilities over
all valid positions.

At the final threshold $p_{\mathrm{true}}\leq0.99$, the six-domain mean
fraction of retained valid positions is 51.1\% for LLaDA and 52.4\%
for Dream.
The retained fraction varies substantially across domains.
For example, only 23.7\% of LLaDA GitHub positions and 32.9\% of
Dream GitHub positions remain below $0.99$, whereas more than 60\%
are retained for HackerNews and Pile-CC.

\begin{figure*}[!t]
    \centering
    \includegraphics[width=0.96\textwidth]
    {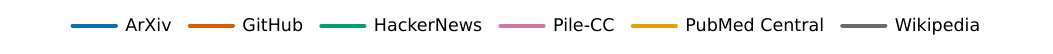}
    \includegraphics[width=0.96\textwidth]
    {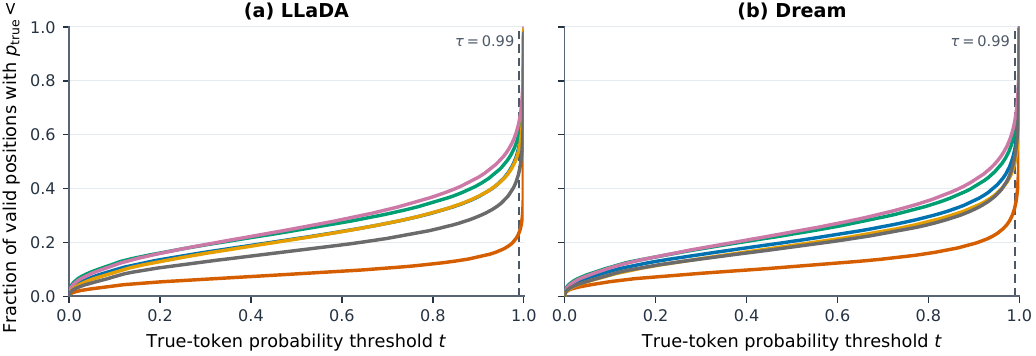}
    \vspace{1mm}
    \caption{
        Empirical CDFs of grouped true-token reconstruction probabilities
        for LLaDA (left) and Dream (right).
        A substantial fraction of valid positions lies in the
        near-deterministic region close to $p_{\mathrm{true}}=1$,
        while the size of this region varies considerably across domains.
    }
    \label{fig:target_only_prob_cdf}
\end{figure*}

The filter therefore does not impose a fixed token budget.
Instead, it removes the portion of each document for which the
reconstruction distribution has already become nearly deterministic.
The CDFs in \cref{fig:target_only_prob_cdf} should not be interpreted
as showing that $0.99$ is mathematically optimal.
Rather, they show that the saturated regime is substantial and that
its prevalence varies across domains.

\subsection{Saturation-Threshold Ablation}
\label[appendix]{app:target_only_threshold_ablation}

We next vary the probability threshold while holding the grouped
measurements and score definitions fixed.
We compare using all valid positions with
$p_{\mathrm{true}}\leq0.50$, $0.90$, $0.95$, and $0.99$.
For each retained support, Masked Score averages $q$, whereas
Target-Only JUMP averages $q+H$.

\begin{table}[H]
\centering
\small
\setlength{\tabcolsep}{3.2pt}
\caption{Mean ROC-AUC across six domains under different near-deterministic filtering thresholds. ``All'' uses every valid position.}
\label{tab:target_only_threshold_ablation}
\begin{tabular}{llccccc}
\toprule
Model & Score & All & $p\leq.50$ & $p\leq.90$ & $p\leq.95$ & $p\leq.99$ \\
\midrule
LLaDA & Masked Score      & .5752 & .5688 & .5858 & .5866 & .5854 \\
LLaDA & Target-Only JUMP  & .6038 & .5785 & .6052 & .6080 & \textbf{.6095} \\
Dream & Masked Score      & .6248 & .6019 & .6393 & .6446 & \textbf{.6454} \\
Dream & Target-Only JUMP  & .6266 & .6015 & .6348 & \textbf{.6377} & .6375 \\
\bottomrule
\end{tabular}
\end{table}

\begin{table}[H]
\centering
\small
\setlength{\tabcolsep}{4pt}
\caption{Mean number of retained positions per document under different probability thresholds, averaged equally across the six domains.}
\label{tab:target_only_retained_counts}
\begin{tabular}{lccccc}
\toprule
Model & All & $p\leq.50$ & $p\leq.90$ & $p\leq.95$ & $p\leq.99$ \\
\midrule
LLaDA & 315.0 & 59.6 & 102.8 & 118.9 & 155.3 \\
Dream & 305.9 & 57.4 & 100.9 & 117.4 & 156.3 \\
\bottomrule
\end{tabular}
\end{table}

\begin{figure*}[!t]
    \centering
    \begin{minipage}{0.48\textwidth}
        \centering
        \includegraphics[width=\linewidth]
        {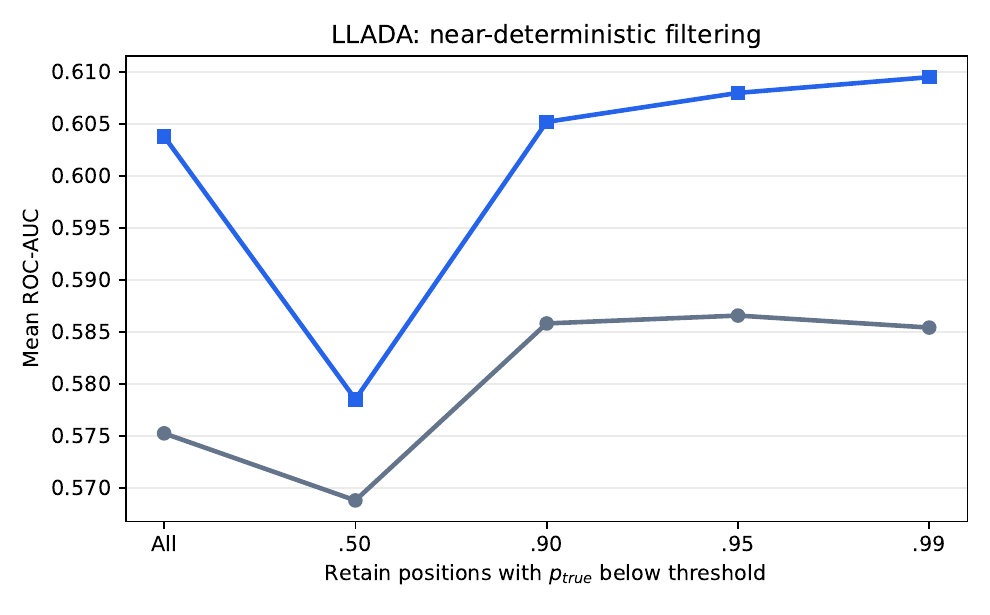}
    \end{minipage}
    \hfill
    \begin{minipage}{0.48\textwidth}
        \centering
        \includegraphics[width=\linewidth]
        {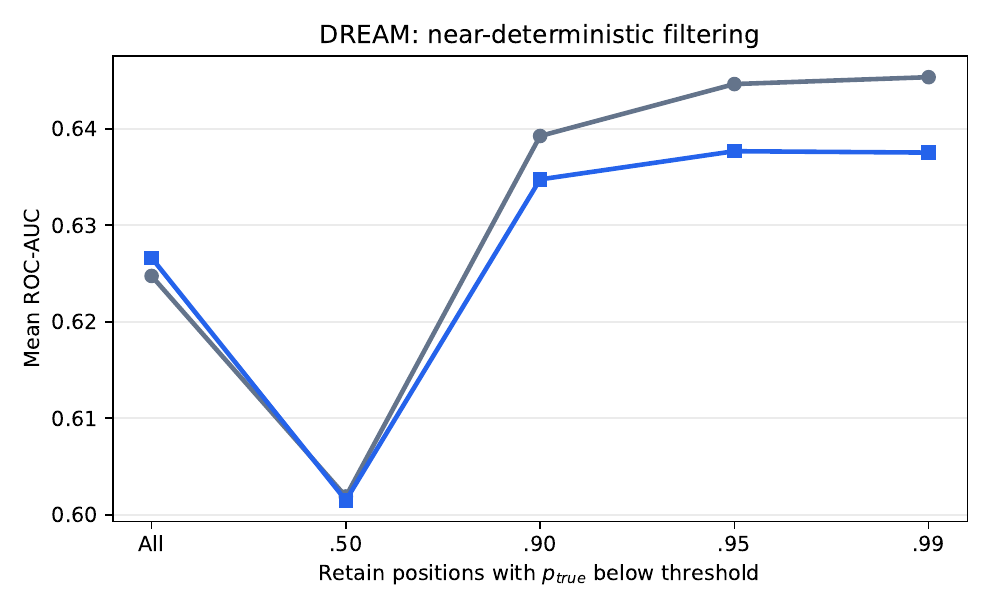}
    \end{minipage}

    \vspace{1mm}
    \includegraphics[width=0.55\textwidth]
    {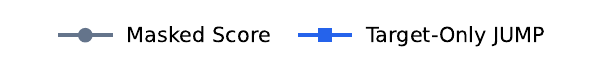}

    \caption{
        Mean ROC-AUC across six domains as the near-deterministic
        filtering threshold varies for LLaDA (left) and Dream (right).
        Aggressive filtering at $p_{\mathrm{true}}\leq0.50$ removes useful
        evidence, whereas performance is comparatively stable over the
        $0.90$--$0.99$ range.
    }
    \label{fig:target_only_threshold_auc}
\end{figure*}

The threshold sweep shows that Target-Only JUMP is not simply a
hard-token selection rule.
The aggressive $p_{\mathrm{true}}\leq0.50$ filter retains only about
60 positions per document on average and reduces attack performance
on both model families.
Performance recovers once the threshold reaches $0.90$ and changes
only modestly over the $0.90$--$0.99$ range.

For LLaDA, the mean ROC-AUC of $q+H$ increases from 0.6052 at
$p_{\mathrm{true}}\leq0.90$ to 0.6080 at $0.95$ and 0.6095 at $0.99$.
For Dream, the corresponding values are 0.6348, 0.6377, and 0.6375.
Thus, $0.99$ is not selected because it uniquely maximizes ROC-AUC.
We use it as a common boundary that removes only the
near-deterministic tail while retaining the broader set of
non-saturated reconstruction events.

\subsection{Membership Signal Inside and Outside the Saturated Regime}
\label[appendix]{app:target_only_saturation_signal}
\begin{figure*}[!t]
    \centering
    \includegraphics[width=0.88\textwidth]
    {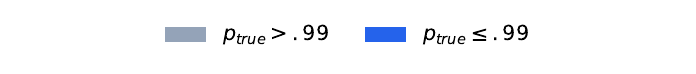}
    \includegraphics[width=0.88\textwidth]
    {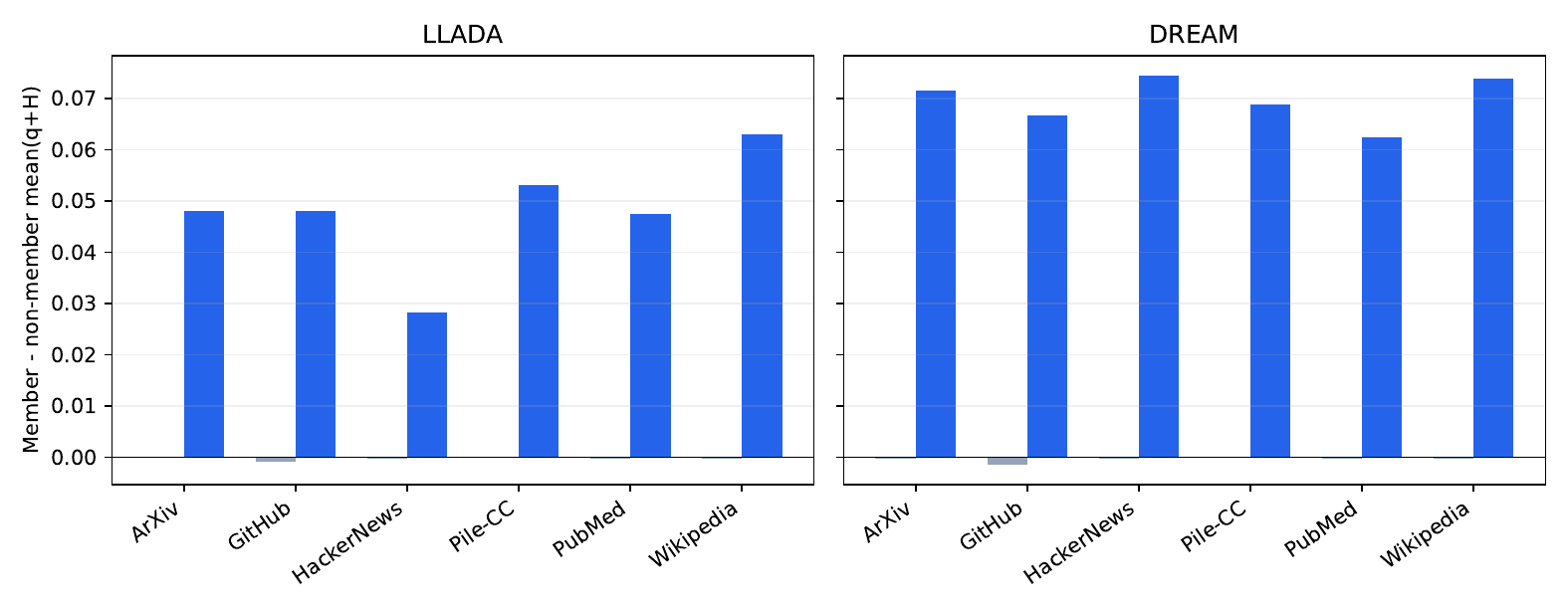}

    \caption{
        Token-level member--non-member difference in $q+H$ inside and
        outside the saturated reconstruction regime.
        Saturated positions contribute essentially no directional
        membership signal, whereas non-saturated positions retain a
        consistently positive member--non-member difference.
    }
    \label{fig:target_only_saturation_gap}
\end{figure*}
The threshold sweep shows that excluding near-deterministic positions
can improve the document-level statistic, but it does not by itself
show whether the removed positions contain membership information.
We therefore directly compare the token-level member--non-member
difference in $q+H$ inside and outside the
$p_{\mathrm{true}}=0.99$ boundary.

\begin{table}[H]
\centering
\small
\setlength{\tabcolsep}{4pt}
\caption{Token-level member--non-member difference in $q+H$ inside and outside the saturated regime. Saturated positions satisfy $p_{\mathrm{true}}>0.99$.}
\label{tab:target_only_saturation_signal}
\begin{tabular}{llccc}
\toprule
Model & Domain & Saturated gap & Non-saturated gap & Saturated fraction \\
\midrule
LLaDA & ArXiv          &  0.00018 & 0.04816 & 44.8\% \\
LLaDA & GitHub         & -0.00090 & 0.04816 & 76.6\% \\
LLaDA & HackerNews     & -0.00002 & 0.02818 & 40.1\% \\
LLaDA & Pile-CC        &  0.00003 & 0.05321 & 35.7\% \\
LLaDA & PubMed Central & -0.00004 & 0.04752 & 46.0\% \\
LLaDA & Wikipedia (en) & -0.00021 & 0.06306 & 53.9\% \\
LLaDA & Mean           & -0.00016 & 0.04805 & 49.5\% \\
\midrule
Dream & ArXiv          & -0.00001 & 0.07149 & 46.2\% \\
Dream & GitHub         & -0.00149 & 0.06670 & 67.2\% \\
Dream & HackerNews     & -0.00029 & 0.07456 & 39.7\% \\
Dream & Pile-CC        &  0.00018 & 0.06890 & 35.7\% \\
Dream & PubMed Central & -0.00014 & 0.06239 & 51.6\% \\
Dream & Wikipedia (en) & -0.00037 & 0.07392 & 48.5\% \\
Dream & Mean           & -0.00035 & 0.06966 & 48.2\% \\
\bottomrule
\end{tabular}
\end{table}

The difference is effectively zero in the saturated region for both
architectures.
Averaged across domains, the member--non-member gap is $-0.00016$
on LLaDA and $-0.00035$ on Dream for positions with
$p_{\mathrm{true}}>0.99$.
In contrast, the corresponding non-saturated gaps are positive:
0.04805 on LLaDA and 0.06966 on Dream.

This behavior matches the limiting argument in the main text.
As $p_{\mathrm{true}}\rightarrow1$, the true-token log-probability
$q$ approaches zero and the predictive entropy $H$ also vanishes.
Consequently, $q+H$ collapses toward zero regardless of membership.
Including many such positions in the document-level average therefore
dilutes evidence from reconstruction events for which multiple
alternatives remain plausible.

\subsection{Entropy-Centered Readout Across Domains}
\label[appendix]{app:target_only_readout_ablation}

We next isolate the contribution of the entropy-centered readout.
Masked Score and Target-Only JUMP use exactly the same
$p_{\mathrm{true}}\leq0.99$ support and the same grouped target outputs.
The only difference is whether the document score averages $q$ or
$q+H$.

\begin{table}[H]
\centering
\small
\setlength{\tabcolsep}{4pt}
\caption{Per-domain ROC-AUC using the same $p_{\mathrm{true}}\leq0.99$ support. Masked Score averages $q$, while Target-Only JUMP averages $q+H$.}
\label{tab:target_only_readout_domains}
\begin{tabular}{lcccc}
\toprule
& \multicolumn{2}{c}{LLaDA} & \multicolumn{2}{c}{Dream} \\
\cmidrule(lr){2-3}\cmidrule(lr){4-5}
Domain & $q$ & $q+H$ & $q$ & $q+H$ \\
\midrule
ArXiv          & .6018 & \textbf{.6169} & .6576 & \textbf{.6592} \\
GitHub         & .5769 & \textbf{.5955} & \textbf{.6372} & .5810 \\
HackerNews     & .5409 & \textbf{.5662} & .6528 & \textbf{.6574} \\
PubMed Central & .5982 & \textbf{.6212} & .6331 & \textbf{.6362} \\
Wikipedia (en) & .6037 & \textbf{.6382} & .6474 & \textbf{.6526} \\
Pile-CC        & .5909 & \textbf{.6188} & \textbf{.6440} & .6388 \\
\midrule
Mean           & .5854 & \textbf{.6095} & \textbf{.6454} & .6375 \\
\bottomrule
\end{tabular}
\end{table}

On LLaDA, entropy centering improves ROC-AUC in all six domains,
raising the six-domain mean from 0.5854 to 0.6095.

The Dream result is less uniform.
Masked Score has a higher six-domain mean ROC-AUC
(0.6454 versus 0.6375), but this difference is dominated by GitHub.
On Dream GitHub, adding entropy decreases ROC-AUC from 0.6372 to
0.5810, a drop of 0.0562.
This single-domain difference is substantially larger than the
differences observed elsewhere.

Across the other five Dream domains, $q+H$ improves ROC-AUC in four.
Pile-CC is the only additional decrease, and its change is much
smaller, from 0.6440 to 0.6388.
As a diagnostic for identifying the source of the Dream mean
difference, excluding GitHub gives a five-domain mean of 0.6470 for
$q$ and 0.6488 for $q+H$.
We do not use this five-domain quantity as an evaluation result; it
only shows that the overall Dream reversal is concentrated primarily
in GitHub rather than reflecting a consistent degradation across
Dream domains.

We retain $q+H$ as the Target-Only JUMP readout because it implements
the within-position relative-preference comparison developed in the
main text and produces consistent improvements across all six LLaDA
domains.
At the same time, Dream GitHub provides a clear counterexample to
uniform dominance.
We therefore report this architecture- and domain-dependent behavior
explicitly rather than claiming that entropy centering always
outperforms masked log-probability.

\subsection{Variance-Normalized Relative Preference}
\label[appendix]{app:target_only_sigma_normalization}

We additionally test whether the relative-preference score benefits
from normalization by the scale of the predictive log-probability
distribution.
Following a Min-K\%++-style normalization, we consider
\begin{equation*}
    z_i
    =
    \frac{q_i+H_i}{\sigma_i},
\end{equation*}
where $\sigma_i$ is the full-vocabulary standard deviation of
log-probabilities under the same grouped reconstruction context.

\begin{table}[H]
\centering
\small
\setlength{\tabcolsep}{4pt}
\caption{Mean ROC-AUC across six domains with and without vocabulary-scale normalization.}
\label{tab:target_only_sigma_normalization}
\begin{tabular}{llccc}
\toprule
Model & Threshold & $q$ & $q+H$ & $(q+H)/\sigma$ \\
\midrule
LLaDA & $p\leq.50$ & .5688 & .5785 & .5634 \\
LLaDA & $p\leq.90$ & .5858 & .6052 & .5923 \\
LLaDA & $p\leq.95$ & .5866 & .6080 & .5962 \\
LLaDA & $p\leq.99$ & .5854 & \textbf{.6095} & .5994 \\
\midrule
Dream & $p\leq.50$ & .6019 & .6015 & .5739 \\
Dream & $p\leq.90$ & .6393 & .6348 & .6215 \\
Dream & $p\leq.95$ & .6446 & .6377 & .6279 \\
Dream & $p\leq.99$ & \textbf{.6454} & .6375 & .6336 \\
\bottomrule
\end{tabular}
\end{table}

\begin{figure*}[!t]
    \centering
    \includegraphics[width=0.88\textwidth]
    {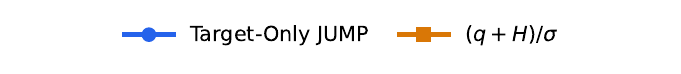}
    \includegraphics[width=0.88\textwidth]
    {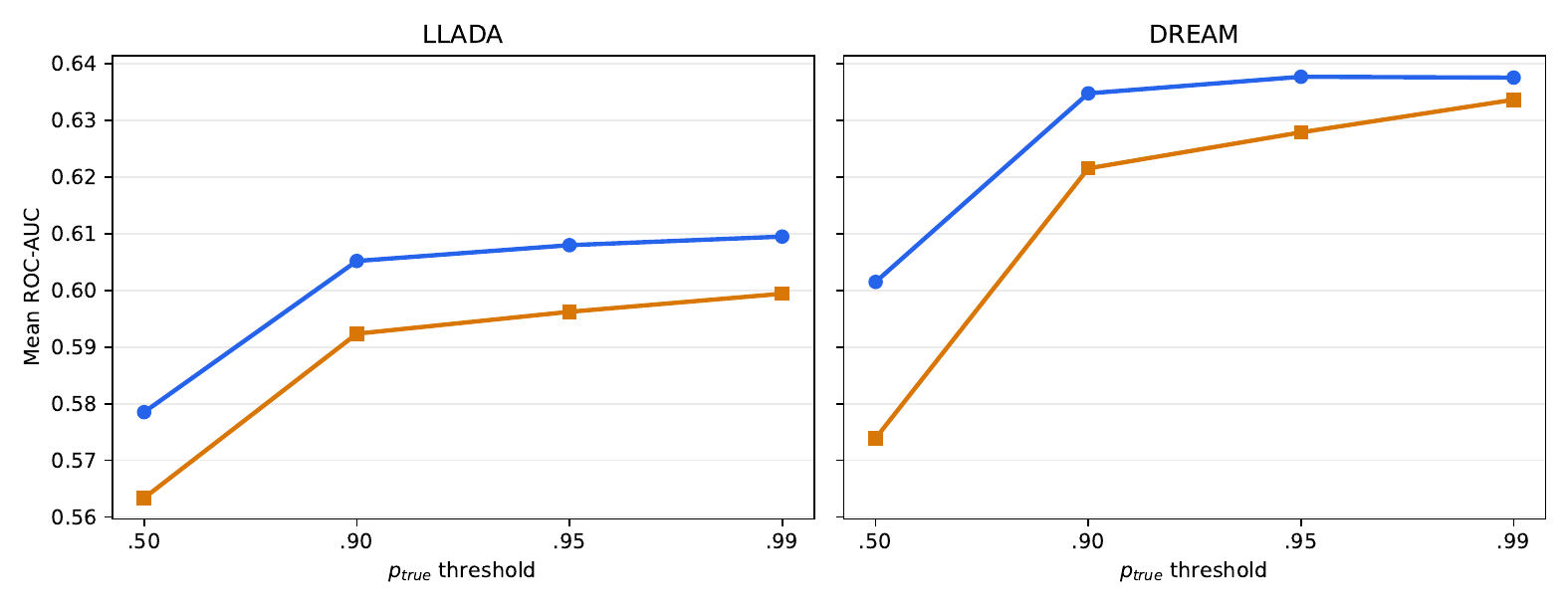}

    \caption{
        Mean ROC-AUC across six domains for the raw masked score,
        entropy-centered relative preference, and its
        vocabulary-scale-normalized variant across probability
        thresholds.
    }
    \label{fig:target_only_sigma_normalization}
\end{figure*}

At the final $p_{\mathrm{true}}\leq0.99$ threshold, normalization reduces
mean ROC-AUC from 0.6095 to 0.5994 on LLaDA and from 0.6375 to
0.6336 on Dream.

Normalization does change individual low-FPR operating points.
On LLaDA, TPR@10\%, TPR@1\%, and TPR@0.1\% FPR change from
0.1995/0.0270/0.0060 for $q+H$ to
0.2060/0.0330/0.0073 after normalization.
On Dream, the corresponding values change from
0.1997/0.0262/0.0053 to
0.2290/0.0300/0.0027.

Thus, vocabulary-scale normalization can alter tail behavior, but it
does not provide a consistent cross-family improvement and reduces
mean ROC-AUC on both model families.
We therefore use the unnormalized $q+H$ readout in the final method.

\subsection{Grouped Approximation to One-Hole Reconstruction}
\label[appendix]{app:target_only_grouped_approximation}

The target-only formulation is motivated by one-hole reconstruction,
whereas the practical attack evaluates positions through a balanced
16-way partition.
We finally test whether the grouped measurement is sufficiently close
to the one-hole view.

On the cached ArXiv diagnostic subset, grouped F16 and exact one-hole
$q$ have a Spearman correlation of 0.9511.
More directly for the final filtering rule, the two measurements agree
on whether a position satisfies $p_{\mathrm{true}}\leq0.99$ for 99.67\%
of the cached positions.
Thus, the grouped approximation closely preserves both reconstruction
confidence ordering and the saturated/non-saturated decision.

We also compare one 16-way partition with two independently hashed
16-way partitions and one 32-way partition.
The latter two settings double the number of masked reconstruction
contexts.

\begin{table}[t]
\centering
\small
\setlength{\tabcolsep}{4pt}
\caption{Grouped reconstruction diagnostic on LLaDA. Values are six-domain means under the final $p_{\mathrm{true}}\leq0.99$ Target-Only JUMP score.}
\label{tab:target_only_grouped_approximation}
\begin{tabular}{lccccc}
\toprule
Measurement & Contexts & AUC & T@10 & T@1 & T@0.1 \\
\midrule
F16 once  & 16 & .6095 & .1995 & .0270 & .0060 \\
F16 twice & 32 & .6130 & .1952 & .0295 & .0062 \\
F32 once  & 32 & .6056 & .1915 & .0255 & .0068 \\
\bottomrule
\end{tabular}
\end{table}

\begin{figure}[htbp]
    \centering
    \includegraphics[width=\columnwidth]
    {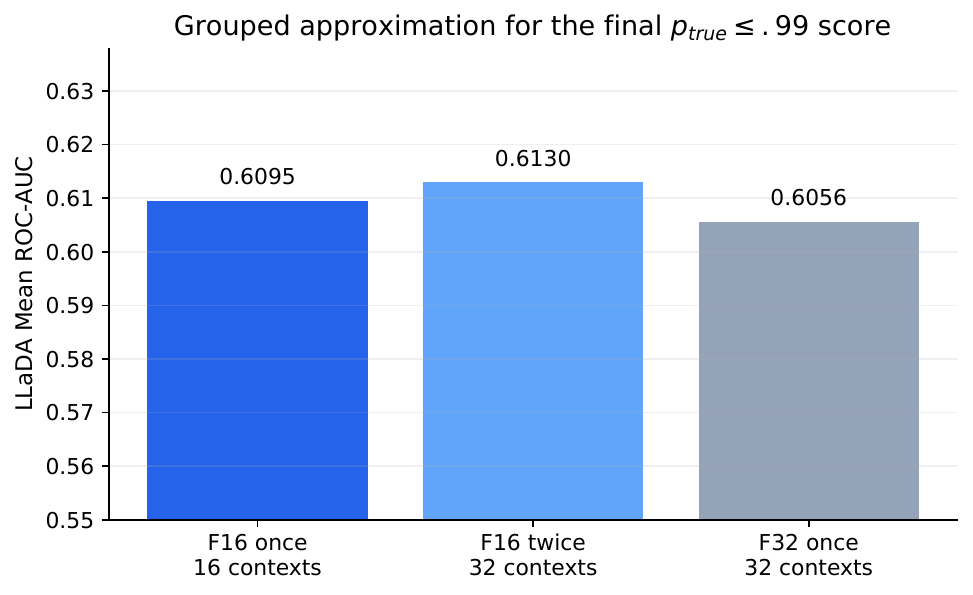}
    \caption{
        Quality--cost comparison for grouped reconstruction under the
        final $p_{\mathrm{true}}\leq0.99$ Target-Only JUMP score.
        Doubling the number of masked contexts yields no material
        improvement in the six-domain mean.
    }
    \label{fig:target_only_grouped_approximation}
\end{figure}

Repeating the 16-way partition increases the six-domain mean ROC-AUC
only from 0.6095 to 0.6130 while doubling the number of masked
contexts from 16 to 32, and slightly reduces TPR@10\% FPR.
Using 32 smaller groups instead yields a mean ROC-AUC of 0.6056.
We therefore retain one balanced 16-way partition as the default.

The exact one-hole diagnostic has one limitation.
The cached exact outputs contain the true-token reconstruction
probability but not the full-vocabulary entropy $H$.
Consequently, we can compare exact and grouped $q$ and the
$p_{\mathrm{true}}\leq0.99$ filtering decision, but cannot compute the
complete exact one-hole $q+H$ document score without additional model
inference.
The reported comparison should therefore be interpreted as evidence
for the fidelity of the grouped reconstruction measurement, rather
than as an exact one-hole evaluation of the complete final attack.

\section{Defense via Matched LoRA and DP-LoRA}
\label[appendix]{app:defense_results}
We separate the effect of adapter parameterization from differential privacy
through two controlled ArXiv experiments: a non-DP LoRA capacity sweep,
followed by a matched non-DP LoRA versus DP-LoRA comparison at fixed adapter
capacity. Our DP experiments build on DP-SGD~\citep{abadi2016deep} and prior work on
differentially private language-model fine-tuning~\citep{yu2022differentially,li2022large}. All rows use 1,000 ArXiv members and 1,000 disjoint non-members.
Reference-assisted JUMP uses the paired pre-fine-tuning checkpoint, while
Target-Only JUMP uses the balanced 16-way reconstruction, fixed
$p_{\mathrm{true}}<0.99$ filtering rule, and entropy-centered readout from
\cref{sec:target_only}.

\subsection{LoRA Capacity Control}
\label[appendix]{app:lora_capacity}

All LoRA targets use four epochs, learning rate $5\times10^{-5}$, effective
batch size 48, maximum length 512, dropout 0.05, seed 42, and $\alpha=2r$.
Held-out reconstruction perplexity is measured on the 1,000 non-members.

\begin{table}[ht]
\centering
\small
\setlength{\tabcolsep}{4pt}
\caption{ArXiv LoRA capacity sweep. JUMP denotes the reference-assisted
attack and Target-Only denotes Target-Only JUMP.}
\label{tab:lora_capacity}
\begin{tabular}{lcccccc}
\toprule
$r/\alpha$ & Trainable & PPL & \multicolumn{2}{c}{AUC} & \multicolumn{2}{c}{TPR@1\%} \\
\cmidrule(lr){4-5}\cmidrule(lr){6-7}
& & & JUMP & Target-Only & JUMP & Target-Only \\
\midrule
16/32 & 0.10\% & 5.22 & 0.512 & 0.498 & 0.007 & 0.012 \\
64/128 & 0.42\% & 4.89 & 0.545 & 0.503 & 0.018 & 0.013 \\
256/512 & 1.65\% & 4.71 & 0.588 & 0.512 & 0.026 & 0.013 \\
1024/2048 & 6.28\% & 4.66 & 0.675 & 0.544 & 0.066 & 0.011 \\
Full FT & 100.00\% & 4.74 & 0.938 & 0.617 & 0.539 & 0.033 \\
\bottomrule
\end{tabular}
\end{table}

Leakage increases with adapter capacity while held-out reconstruction improves.
Reference-assisted JUMP rises from $0.512$ to $0.675$ across the LoRA sweep,
and Target-Only JUMP rises from $0.498$ to $0.544$. Full fine-tuning produces
the strongest signal for both attacks, reaching $0.938$ and $0.617$,
respectively. The paired checkpoint therefore reveals a sharper capacity
trend, while the target-only score preserves its direction at higher adapter
capacity.

\subsection{Matched LoRA and DP-LoRA}
\label[appendix]{app:matched_dp_lora}

We next fix $r=1024$, $\alpha=2048$, dropout 0.05, clipping norm 1.0, and
$\delta=10^{-5}$, and vary only the DP noise required for target privacy
budgets $\varepsilon\in\{0.01,1,10,20\}$. The non-DP LoRA row uses the
identical adapter capacity and training data.

\begin{table}[ht]
\centering
\small
\setlength{\tabcolsep}{4pt}
\caption{Matched non-DP LoRA and DP-LoRA on ArXiv across target privacy budgets.}
\label{tab:matched_dp_lora}
\begin{tabular}{lcccccc}
\toprule
Method & Target $\varepsilon$ & PPL & \multicolumn{2}{c}{AUC} & \multicolumn{2}{c}{TPR@1\%} \\
\cmidrule(lr){4-5}\cmidrule(lr){6-7}
& & & JUMP & Target-Only & JUMP & Target-Only \\
\midrule
Non-DP LoRA & N/A & 4.66 & 0.675 & 0.544 & 0.066 & 0.011 \\
DP-LoRA & 0.01 & 55.77 & 0.529 & 0.497 & 0.013 & 0.012 \\
DP-LoRA & 1 & 18.67 & 0.502 & 0.498 & 0.012 & 0.011 \\
DP-LoRA & 10 & 7.89 & 0.514 & 0.498 & 0.013 & 0.011 \\
DP-LoRA & 20 & 7.05 & 0.520 & 0.498 & 0.014 & 0.011 \\
\bottomrule
\end{tabular}
\end{table}

At fixed adapter capacity, DP reduces reference-assisted JUMP from $0.675$ to
between $0.502$ and $0.529$, and Target-Only JUMP from $0.544$ to between
$0.497$ and $0.498$. The target-only values and their TPR@1\% FPR are at
chance-level resolution throughout the DP sweep. The strictest privacy budgets
also increase held-out reconstruction perplexity, while weaker budgets retain
more domain utility.

\paragraph{Accounting caveat.}
Privacy accounting uses a PRV accountant with a fixed-batch sampling-rate
approximation and \texttt{secure\_rng=False}. The realized privacy budgets
closely match the target $\varepsilon$ values reported in
\cref{tab:matched_dp_lora}. These values are accountant outputs under this
implementation, not an independently certified deployment guarantee.
Accordingly, these experiments demonstrate attack suppression and a trade-off
between privacy and utility under a matched training setup; they should not be
read as a production certification.
\FloatBarrier

\Needspace{7\baselineskip}
\section{Existing Assets and Licenses}
\label[appendix]{app:licenses}
We use publicly released models, datasets, and baseline resources.
\cref{tab:asset_licenses} summarizes their roles and the license or access terms used by this work.
We do not redistribute pretrained model weights or the original MIMIR data.
Users should obtain each asset from its original provider and comply with its license and terms of use.

\begin{table}[H]
\centering
\small
\setlength{\tabcolsep}{3pt}
\renewcommand{\arraystretch}{1.15}
\caption{Existing assets used in this paper.}
\label{tab:asset_licenses}
\begin{tabular}{
p{0.18\textwidth}
p{0.30\textwidth}
p{0.20\textwidth}
p{0.24\textwidth}
}
\toprule
Asset & Role in this paper & Citation / source & License / terms \\
\midrule

LLaDA-8B-Base
& Main base/reference dLLM and fine-tuning target
& \citet{nie2025large}
& MIT \\

Dream-v0-Base-7B
& Second dLLM family for cross-architecture evaluation
& \citet{ye2025dream}
& Apache-2.0 \\

MIMIR
& Member/non-member evaluation benchmark
& \citet{duan2024membership}
& MIT; gated access terms on Hugging Face \\

C4
& PRISM selector training corpus
& \citet{raffel2020exploring}
& ODC-BY; Common Crawl Terms of Use \\

SlimPajama
& Selector-corpus ablation
& \citet{soboleva2023slimpajama}
& Apache-2.0 for the original Cerebras release; subset-specific terms if using a split or reupload \\

FineWeb
& Selector-corpus ablation
& \citet{penedo2024fineweb}
& ODC-By v1.0; Common Crawl Terms of Use \\

SAMA
& Prior dLLM MIA baseline
& \citet{chen2026membership}
& MIT if using the official code repository \\

\bottomrule
\end{tabular}
\end{table}

\FloatBarrier

\end{document}